\documentclass[letterpaper, 10 pt, conference]{ieeeconf}  % Comment this line
\IEEEoverridecommandlockouts                              % This command is only needed if
\title{Anytime Game-Theoretic Planning with Active Reasoning About Humans' Latent States for Human-Centered Robots}

\author{Ran Tian, Liting Sun, Masayoshi Tomizuka, and David Isele
 \thanks{
         Ran Tian, Liting Sun, and Masayoshi Tomizuka are with the UC Berkeley.
          ({\tt\small \{rantian,liting,tomizuka\}@berkeley.edu}). David Isele is with Honda Research Institute. ({\tt\small disele@honda-ri.com}).}
}
\vspace{-0.5cm}
\usepackage{tabularx}
\usepackage{float}
\usepackage{epsfig}
\usepackage{epstopdf}
\usepackage{bbm}
\usepackage{mathtools,xparse}

\usepackage{epsfig} % for postscript graphics files
\usepackage{amsmath} % assumes amsmath package installed
\usepackage{amssymb}  % assumes amsmath package installed
\usepackage{bm}
\usepackage{epstopdf}
\usepackage{color}
\usepackage{etoolbox}
\DeclareMathOperator*{\argmax}{arg\,max}

\usepackage{comment}
\usepackage{diagbox}
\usepackage[ruled, lined, longend, linesnumbered, noend]{algorithm2e}
\usepackage{booktabs,multirow}
\usepackage{bigstrut}
\usepackage{cite}
\usepackage{hyperref}
\usepackage[capitalise]{cleveref}

\usepackage{lipsum}
\usepackage{color}
\usepackage[colorinlistoftodos]{todonotes}
\usepackage{lipsum}
\usepackage{hyperref}
\usepackage[nodisplayskipstretch]{setspace}
\usepackage{amsfonts}
\usepackage{mathtools}
\usepackage{graphicx}
\newcommand{\tr}[1]{{\color{black}#1}}

\begin{document}
\maketitle
\begin{abstract}
A human-centered robot needs to reason about the cognitive limitation and potential irrationality of its human partner to achieve seamless interactions. This paper proposes an anytime game-theoretic planner that integrates iterative reasoning models, a partially observable Markov decision process, and chance-constrained Monte-Carlo belief tree search for robot behavioral planning. Our planner enables a robot to \textit{safely} and \textit{actively} reason about its human partner's latent cognitive states (bounded intelligence and irrationality) in real-time to maximize its utility better. We validate our approach in an autonomous driving domain where our behavioral planner and a low-level motion controller hierarchically control an autonomous car to negotiate traffic merges. Simulations and user studies are conducted to show our planner's effectiveness.
\end{abstract}

\vspace{-0.2cm}
\section{Introduction}
%% background
Human-centered robots \tr{(e.g., self-driving cars, assistive robots, etc.)} operate in close proximity to humans. When designing planning algorithms for human-centered robots, it is critical for the robot to reason about the mutual influence between itself and human actors. Such a mutual dependency can be formulated as a general-sum game, in which a standard approach is to assume that each agent is a perfectly rational, expected utility maximizer, who simultaneously responds optimally to all the others (i.e., operates under equilibrium strategies) \cite{myerson2013game, schwarting2019social}. However, experimental studies \cite{goeree2001ten, crawford2007level, wright2014level} suggest that human behaviors often systemically deviate from equilibrium behaviors due to their latent cognitive states: \textit{bounded intelligence} (cognitive limitation) and \textit{irrationality} (tendency to make errors). Therefore, a robot must account for its human partner's cognitive states for seamless and safe interactions.

% recent works 
Recent works exploited the leader-follower model \cite{Sadighinformation,sadigh2016planning,fisac2019hierarchical,stefansson2019human} and the level-$k$ model \cite{LiUnsignalized, tian2018adaptive, Sisi2019} to equip robots with the ability to reason about humans' non-equilibrium behaviors. These planners either assign humans' latent states \textit{a priori}, omitting humans' distinct cognitive characteristics, or passively adapt to the humans' latent states, sacrificing the benefits from actively learning the latent states (Sec.~\ref{sec: related work}).

% our approach
In this work, we propose an anytime game-theoretic planning framework that integrates iterative reasoning models, a partially observable Markov decision process (POMDP), and chance-constrained Monte-Carlo belief tree search. Drawing inspiration from behavioral game theory, we model humans' intelligence levels and degrees of rationality as their latent cognitive states, capturing their heterogeneous cognitive limitations and tendencies to make errors. Rather than passively adapting to humans' latent states when planning, our approach enables the robot to actively and safely learn the latent states to achieve its goal more effectively \tr{without losing the ability of real-time execution}. Our key insight is: \textit{Human-centered robots can exploit the mutual influence in interactions to design actions that reveal their human partners' cognitive limitations and degrees of rationality. By actively reasoning about these latent states, robots can achieve more effective planning.} 
Overall, we make the following contributions:

% \begin{figure}[t]
% \begin{center}
% \begin{picture}(300.0, 40)
% %\put(0, 0){\epsfig{file=media/intro_fig.png,width = 0.55 \linewidth,angle=0, trim=10cm 0.0cm 8.0cm 0.5cm,clip}}
% %\put(135, 5){\epsfig{file=media/pnc.png,width = 0.43 \linewidth, angle=0, trim=0cm 0.0cm 0.0cm 0.0cm,clip}}
% %\put(0, 0){\epsfig{file=media/intro_fig.png,width = 0.6 \linewidth,angle=0, trim=8cm 0.0cm 8.0cm 0.5cm,clip}}
% %\put(45, 5){\epsfig{file=media/intro_fig_2.png,width = 0.4 \linewidth, angle=180, trim=8cm 0.0cm 8.0cm 0.0cm,clip}}
% \put(8, 5){\epsfig{file=media/introl.png,width = 0.9 \linewidth,angle=0, trim=0cm 0.0cm 0.0cm 0.0cm,clip}}
% \small
% \normalsize
% \end{picture}
% \end{center}
% \vspace{-0.8cm}
% \caption{\hspace{-0.3cm} An autonomous car interacts with a human-driven car controlled by a user through a racing wheel and a pedal in a forced merging scenario.}
% %\caption{Left: An autonomous car interacts with a human-driven car. Right: Hierarchical planning \& control for the robot.}
% \label{fig: intro_fig}
% \vspace{-0.6cm}
% \end{figure}

% Our contributions
% \noindent 
% \textbf{Formulation of planning with humans' cognitive states.} We formalize the robot planning problem in a human-robot team as a POMDP. We exploit the quantal level-$k$ model from behavioral game theory to model the human's bounded intelligence and irrationality, which are modeled as latent states in the POMDP. In addition, a value iteration algorithm is proposed to construct closed-loop human behavioral models under various intelligence levels and irrationality.

\noindent
\textbf{An active anytime game-theoretic planner.} We formalize the robot planning problem in a human-robot team as a POMDP with the human's cognitive states as latent states. The POMDP is approximately solved by an open-loop Monte-Carlo belief tree search algorithm in an anytime manner. Coupled with explicit realization of active information gathering on the latent states, chance-constrained branching, and tailored terminal value functions, our planner enables the robot to safely and adaptively balance between exploration (learning the human's latent states) and exploitation (maximizing utility).

\noindent
\textbf{Application of our framework to autonomous driving.} The proposed behavioral planner is connected with an off-the-shelf low-level motion control layer \cite{borrelli2017predictive} to achieve feedback control for an autonomous car represented by a high-fidelity model. Simulations and user studies are conducted to show the effectiveness of our planner compared to baselines.

\vspace{-0.13cm}
\section{Related Work}\label{sec: related work}

\noindent\textbf{Game-theoretic planning for human-centered robots.} \tr{Our work is related to \cite{Sadighinformation, sadigh2016planning, fisac2019hierarchical, stefansson2019human, LiUnsignalized, tian2018adaptive, Sisi2019}. In \cite{sadigh2016planning, fisac2019hierarchical, stefansson2019human}, the robot exploits the Stackelberg game \cite{simaan1973additional} and models its human partner as a \textit{follower} who accommodates the robot's planned actions. The follower model is homogeneous (the human is always the follower w.r.t. the robot), thus the robot may behave poorly when the human does not behave like a follower. The approach in \cite{Sadighinformation} allows the robot to \textit{actively} ``probe" the human's latent states, but the underlying human model is still a pure follower model. In addition, the safety of the planner is not explicitly enforced in \cite{Sadighinformation}. In \cite{LiUnsignalized, tian2018adaptive, Sisi2019}, the robot exploits the level-$k$ model \cite{costa2001cognition} to model the human as an agent who reasons under various intelligence levels. While the level-$k$ model is heterogeneous, it assumes that humans with lower-levels of intelligence best respond to higher-level humans, omitting humans' potential irrationality. Furthermore, the planners in \cite{LiUnsignalized, tian2018adaptive, Sisi2019} \textit{passively} adapt to the human's intelligence level, leading to less effective plans. \textit{Our approach is different from \cite{Sadighinformation, sadigh2016planning, fisac2019hierarchical, stefansson2019human} since our planner does not assume humans' cognitive  characteristics in interactions a \textit{priori} but reasons about them when planning. Our work is also distinguished from \cite{LiUnsignalized, tian2018adaptive, Sisi2019} by enabling the robot to safely and actively learn the human's cognitive states to maximize utility more effectively without losing the ability of real-time execution.}}

% \textit{Here, we embrace that humans have heterogeneous intelligence levels and degrees of rationality as their latent states. Our planner actively learns these latent states to better maximize utility, and handles safety explicitly by enforcing chance constraints.} 

%Recent works also exploited other human models for designing planning algorithms, including: trust \cite{chen2018planning, chen2020trust} and risk models \cite{kwon2020humans, tian2020bounded}. Our work is not competing with these works, but rather is complementing them. %For example, the risk-measure exploited in \cite{kwon2020humans, tian2020bounded} can be plugged in our planning framework seamlessly.

% This part will likely to be cutoff due to page limit, but can be left in the online version
\noindent\textbf{Solution methods for POMDP.} A POMDP is a framework for planning under uncertainty. Various approaches have been proposed to approximate POMDPs, including point-based methods \cite{pineau2003point, porta2006point, kurniawati2008sarsop}, open-loop strategies \cite{yu2005open, weinstein2013open,perez2015open, phan2019memory}, and Monte-Carlo tree search \cite{silver2010monte,lim2019sparse}. Partially observable Monte-Carlo planning (POMCP) \cite{silver2010monte} performs well, though building a closed-loop search tree in large games is computationally expensive. Open-loop strategies condition action selection on previous action sequences. They use a much smaller search space by sacrificing the ability of active information gathering, and achieve competitive performance compared to closed-loop planning under computational constraints \cite{weinstein2013open}. \textit{Our approach combines the strengths from POMCP and open-loop strategies, achieving real-time active game-theoretic planning.}
\vspace{-0.2cm}
\section{Problem formulation}\label{sec: problem formulation}

\noindent
\textbf{Human-robot team formalization.}
We formalize the human-robot team as a two-player dynamic game represented by the tuple $\mathcal{G} = <\mathcal{P}, \tilde{\mathcal{S}}, \mathcal{A}, f, r_{\mathcal{R}}, r_\mathcal{H}>$, where $\mathcal{P} = \{\mathcal{R},\mathcal{H}\}$ represents the two players with $\mathcal{R}$ denoting the robot and $\mathcal{H}$ denoting the human; $\tilde{\mathcal{S}} = \tilde{\mathcal{S}}_{\mathcal{R}}\times \tilde{\mathcal{S}}_{\mathcal{H}}$ and $\mathcal{A} = \mathcal{A}_{\mathcal{R}} \times \mathcal{A}_{\mathcal{H}}$ are, respectively, the joint fully-observable state and action spaces of the two agents; the function $f$ governs the evolution of the joint fully-observable state and is defined by the following dynamic model: $\tilde{s}_{t+1} = f(\tilde{s}_t,a^{\mathcal{R}}_t, a^{\mathcal{H}}_t)$ ( $\tilde{s}_t \in \tilde{\mathcal{S}}$, $a^{\mathcal{R}}_t\in\mathcal{A}_{\mathcal{R}}$, $a^{\mathcal{H}}_t\in\mathcal{A}_{\mathcal{H}}$); $r_{(\cdot)}: \tilde{\mathcal{S}} \rightarrow \mathbb{R}$ is the reward function of an agent.%s, with $r_{(\cdot)}: \tilde{\mathcal{S}} \rightarrow \mathbb{R}$ representing their planning goals. 

\noindent
\textbf{Robot planning as a POMDP.} We consider planning from the robot's perspective. The fully-observable state $\tilde{s}$ of the human-robot team  represents measurable variables (e.g., position, speed, etc.). In addition, the human also has latent states (e.g., preference, trust, cognitive limitation, etc.) that characterize his/her cognition and reasoning; such latent states cannot be observed directly, and therefore must be inferred from interactions. We let $\theta \in \Theta$ denote the human's latent states, and we consider an augmented state space $\mathcal{S} = \tilde{\mathcal{S}} \times \Theta$. As the robot's knowledge about the augmented state $s \in \mathcal{S}$ is incomplete, it maintains a belief distribution over $\mathcal{S}$ at each discrete time step $t$, namely, the robot maintains the belief state $b_t := [\mathbb{P}(s_t = s_1),\dots,\mathbb{P}(s_t = s_{|\mathcal{S}|})]^{\intercal}$. We formulate the robot planning problem as a POMDP defined by the tuple $<\mathcal{G}, \mathcal{S}, \mathcal{B}, \Omega, \mathcal{Z}, \rho, r'_{\mathcal{R}}, \mathbb{O}_{\text{safe}}>$, where $\mathcal{G}$ denotes the dynamic game model defined above; $\mathcal{S}$ is the augmented state space; $\mathcal{B}$ represents the space of probability distributions over $\mathcal{S}$ ($b_t\in\mathcal{B}$); $\Omega$ is the finite observation space; \small $\mathcal{Z}: \Omega \times \mathcal{S} \rightarrow [0,1]$ \normalsize is a probability function specifying the probability of receiving an observation in a state; the belief dynamics function $\rho: \mathcal{B} \times \mathcal{A}_{\mathcal{R}} \times \Omega\rightarrow \mathcal{B}$ governs the belief state evolution and is defined as $b_{t+1} = \rho(b_t,a^{\mathcal{R}}_t,o_{t+1})$. Given an initial belief state $b_t$, the robot executes the action $a^{\mathcal{R}}_t$, receives the observation $o_{t+1}$ at time step $t+1$, and updates its belief accordingly. $r'_{\mathcal{R}}: \mathcal{B} \times \mathcal{A}_{\mathcal{R}} \rightarrow \mathbb{R}$ denotes the reward function of the robot in belief space (defined in Sec.~\ref{sec: embed human model}); $\mathbb{O}_{\text{safe}}\subseteq \tilde{\mathcal{S}}$ represents the set of safe states of the robot.

We let $\pi_{\mathcal{R}}: \mathcal{B} \rightarrow \mathcal{A}_{\mathcal{R}}$ denote a robot's deterministic policy. Given belief state $b_t$, the robot maximizes its value: 
\vspace{-0.2cm}
\begin{align}\label{equ: closed loop policy}
% \hspace{+5cm} 
\pi^{*}_{\mathcal{R}} = \argmax_{\pi} V_{\mathcal{R}}^{\pi}(b_t),
\end{align}
\normalsize
\vspace{-0.45cm}

\noindent
where \small $\small V_{\mathcal{R}}^{\pi}(b_t) = \mathbb{E}_{\mathcal{Z}} \left[\sum_{\tau=0}^{\infty} \gamma^\tau r'_\mathcal{R}(b_{t+\tau}, a_{t+\tau}) \big| a_{t+\tau}=\pi(b_{t+\tau})\right]\normalsize$ \normalsize is the value function representing the robot's expected return starting from $b_t$, subject to its policy and the belief dynamics function.
% ; the expectation is taken with respect to the possible future observations. 
Note the robot needs to reason about both its own actions and its human partner's responses due to the mutual dependence. The POMDP formulation allows us to condition the robot behaviors on the inferred latent states.% as the reasoning of human behaviors can be embedded in the belief dynamics function.
%\di{Note: this description is good, but a bit long. If there are space issues later, we can shorten it.}\tr{shorted a bit}

%\noindent
%\textbf{Tasks.} In what follows, we present two efforts in solving for $\pi_{\mathcal{R}}$: 1) building the human computational model and embed that in the POMDP; 2) approximating the solution.

% In this work, we draw inspiration from  behavioral game theory to model the human's latent state as his/her \textit{intelligence level} and \textit{rationality coefficient}, capturing the human's heterogeneous cognitive limitations and degrees of rationality. Moreover, instead of computing infinite horizon closed-loop policies, we integrate open-loop control in POMCP, %. Coupled with explicit realization of active information gathering on the human’s latent state, tailored  terminal  state  value  functions,  and  pruning  search nodes that violate the chance-constraint of safety, our planner
%enabling the robot to safely and adaptively balance the \textit{exploration} (revealing the human's latent state) and the \textit{exploitation} (executing the robot's nominal objective). 

\vspace{-0.2cm}
\section{Human latent states modeling}\label{sec: internal state modeling}

\subsection{Iterative Reasoning Model}

When studying interactions in games, players are commonly assumed to adopt the Nash equilibrium solution: each player has unlimited computational resources and responds optimally to the others. 
In real life however, humans are known to act irrationally and have cognitive limitations. The iterative reasoning models from behavioral game theory have been proven to show better performance in characterizing humans' reasoning capabilities in simultaneous games \cite{wright2014level}. Examples of iterative reasoning models include: the level-$k$ model \cite{costa2001cognition}, the cognitive hierarchy model \cite{camerer2004cognitive}, and the quantal level-$k$ model \cite{stahl1994experimental}. All these models aim to capture humans' cognitive limitations and share a common feature: they model humans as agents with heterogeneous \textit{bounds} on their reasoning abilities, i.e., human agents can only perform a finite number of iterations of reasoning, and such an intelligence bound is referred as the \textit{intelligence level}. Among the various integrative reasoning models, the quantal level-$k$ model is the state-of-the-art \cite{WRIGHT201716}.

%The level-$k$ framework has been exploited for modeling human interactions in autonomous driving applications in our previous works \cite{tian2018adaptive, tian2019game}. We note that the key difference between the quantal level-$k$ framework and the level-$k$ framework is the error structure. In the level-$k$ framework, higher-level agents assume the best responses from and best respond to lower-level agents, although human agents are likely to make sub-optimal actions and errors. In contrast, in the quantal level-$k$ framework, higher-level agents are aware that lower-level agents have some probability of making sub-optimal actions and errors. The policies in the level-$k$ framework are set-valued functions, while the quantal level-$k$ policies are mixed strategies.

\vspace{-0.2cm}
\subsection{Human Quantal Level-k Reasoning Model}
\label{sec: human qlk reasoning}

\noindent
\textbf{Quantal best response and rationality coefficient.} One of the key components of the quantal level-$k$ (ql-$k$) model is quantal best response. The notion behind quantal best response is that human players are more likely to select actions with higher expected future rewards \cite{mckelvey1995quantal}. Formally, we define the quantal best response function as follows: let $Q^{i}(\tilde{s},a^i|a^{-i})$ denote agent $i$'s expected total reward ($i\in\mathcal{P}$) when executing $a^{i}$ in $\tilde{s}$ against an action $a^{-i}$ from his/her opponent $-i$. % ($a^{-i}$ denotes agent $i$'s opponent's action).
Then a quantal best response by agent $i$ to agent $-i$ is a mixed policy:

\vspace{-0.4cm}
\small
\begin{align}\label{equ: quantal best response}
\hspace{+0.5cm}\pi^{i}(\tilde{s},a^i|a^{-i}) = \frac{\text{exp}\big( \lambda^i Q^{i}(\tilde{s},a|a^{-i})\big)}{\sum_{a'\in \tilde{\mathcal{A}}_{i}} \text{exp}\big(\lambda^i Q^{i}(\tilde{s},a'|a^{-i})\big)} \enspace,
\end{align}
\vspace{-0.3cm}
\normalsize

\noindent where $\lambda^i \in (0,1]$ is the \textit{rationality coefficient} that controls the degree of agent $i$ conforming to optimal behaviors. In general, the larger the $\lambda$ is, the more rational the human is.%, and the less likely that the human will make errors when making decisions.

% Add one sentense to address the releation between QBS and qlk

%The quantal level-$k$ model combines the notations of quantal best response and iterative reasoning.

\noindent
\textbf{Human quantal level-$k$ policies.} In the ql-$k$ model, the iterative reasoning process starts from ql-$0$ agents who are
non-strategic reasoners. %\footnote{Strategic reasoners model other agents' potential responses when making decisions while non-strategic reasoners do not.}.
Then, a ql-$k$ agent, $k\in\mathbb{N}^+$, assumes the other agents are ql-$(k-1)$ agents, predicts their ql-$(k-1)$ policies, and quantally best responds to the predicted ql-$(k-1)$ policies. On the basis of ql-$0$ policies, the ql-$k$ policies are defined for every $i\in\mathcal{P}$, for every $\lambda\in\Lambda$, and for every $k = 1,\dots,k_{\text{max}}$ through a sequential and iterative process. Specifically, given an initial state $\tilde{s}_t \in \tilde{\mathcal{S}}$, a ql-$k$ agent $i$ maximizes the following objective: $\max_{\pi^{i,k,\lambda^i}} V^{i,k}(\tilde{s}_t)$, where \small $V^{i,k}(\tilde{s}_t)= \mathbb{E}_{\pi^{*,-i,k-1,\lambda^{-i}}}\Big[ \sum_{\tau=0}^{\infty}\gamma^\tau r_{i}(\tilde{s}_{t+\tau})\Big]$ \normalsize is the ql-$k$ value function of agent $i$ and \small $\pi^{*,-i,k-1,\lambda^{-i}}: \tilde{\mathcal{S}}\times\mathcal{A}_{-i} \rightarrow [0,1]$ \normalsize is the predicted ql-$(k-1)$ policy of agent $-i$. The optimal value function satisfies the following Bellman equation: \footnotesize $ V^{*,i,k}(\tilde{s}) = \mathcal{B}\ V^{*,i,k}(\tilde{s})=\max_{a^i\in\mathcal{A}_i} \mathbb{E}_{\pi^{*,-i,k-1,\lambda^{-i}}}\Big[r_{i}(\tilde{s}') + \gamma V^{*,i,k}(\tilde{s}') \big |\tilde{s}' = f(\tilde{s},a^i,a^{-i}), a^{-i} \sim \pi^{*,-i,k-1,\lambda^{-i}}\Big]$ \normalsize, and can be determined via value iteration. Then, we define the $Q$-value function as:

\vspace{-0.4cm}
\small
\begin{align}\label{equ: quantal Q-value}
    Q^{*,i,k} (\tilde{s},a^i)= \mathbb{E}_{\pi^{*,-i,k-1,\lambda^{-i}}} \big[r_{i}(\tilde{s}) + \gamma V^{*,i,k}(\tilde{s}')\Big],
\end{align}
\normalsize
\vspace{-0.5cm}

\noindent and \eqref{equ: quantal best response} is adopted to define agent $i$'s ql-$k$ policy. We note that when agent $i$ predicts its opponent's ql-$(k-1)$ policy, it assumes that its opponent's rationality coefficients is also $\lambda^i$ (i.e., $\lambda^{-i} = \lambda^i$) and forms its policy based on $\pi^{*,-i,k-1,\lambda^i}$. We summarize the algorithm that computes the ql-$k$ policies of the agents in $\mathcal{G}$ in Alg.~\ref{alg: quantal level-k dp}. %Note that: 1) Alg.~\ref{alg: quantal level-k dp} exploits a closed-loop feedback information structure to compute game-theoretic policies, as opposed to the open-loop information structure used in \cite{spica2020real, tian2018adaptive}; 2) the algorithm extends to multi-player dynamic games straightforwardly.

\vspace{-0.4cm}
\begin{algorithm}
    \footnotesize
    \caption{Quantal level-$k$ dynamic programming}
    \label{alg: quantal level-k dp}
    
    \textbf{Input}: The highest intelligence level $k_{\text{max}}$, a set of rationality coefficients $\Lambda$, and the level-$0$ model $\pi^{i,0}$, $i\in\mathcal{P}$.
    
    %\textbf{Output}: $\{\pi^{*,i,k,\lambda}\}$, $i\in\mathcal{P}$, $k =1,\dots,k_{\text{max}}$, and $\lambda \in \Lambda$.
    
    \For{$k = 1: k_{\text{max}}$}{
        \For{$(i,\lambda) \in \mathcal{P}\times\Lambda$}{
        
               % Initialize $V^{i,k}(\tilde{s}), \forall \tilde{s} \in \tilde{\mathcal{S}}$;
            
            \While{$V^{i,k}$ not converged}{
                
                \textbf{for} $\tilde{s}\in\tilde{\mathcal{S}}$ \textbf{do} $V^{i,k}(\tilde{s}) \leftarrow \mathcal{B} V^{i,k}(\tilde{s})$;
                
                % \For{$\tilde{s}\in\tilde{\mathcal{S}}$}{
                %     $V^{i,k}(\tilde{s}) \leftarrow \mathcal{B} V^{i,k}(\tilde{s})$;
                % }
            }
            \For{$(\tilde{s},a^i) \in \tilde{\mathcal{S}}\times \mathcal{A}_i$}{
                
                %$ Q^{*,i,k} (\tilde{s},a^i)= \mathbb{E}_{\Pi^{*,-i,k,\lambda}} \big[r_{i}(\tilde{s}) + \gamma V^{i,k}(\tilde{s}')\Big], \tilde{s}' = f(\tilde{s},a^i,a^{-i});$
                Compute $\pi^{*,i,k,\lambda} (\tilde{s}, a^i) =\hspace{-0.1cm} \frac{\text{exp}\big( \lambda Q^{*,i,k}(\tilde{s},a^i)\big)}{\sum_{a'\in \mathcal{A}_{i}} \text{exp}\big(\lambda Q^{*,i,k}(\tilde{s},a')\big)}$ using \eqref{equ: quantal Q-value};
            }
        }
    }
    Return $\{\pi^{*,i,k,\lambda}\}$, $i\in\mathcal{P}$, $k = 1,\dots,k_{\text{max}}$, and $\lambda\in\Lambda$.
    \normalsize
\end{algorithm}
% \vspace{-0.5cm}
%\di{the way line 13 is currently written, there is no dependence on $a$}\tr{fixed}
\vspace{-0.4cm}

% put single-shot game... to the next section 

\noindent
% \textbf{Summary.} We model the human's latent states as his/her intelligence level and rationality coefficient, i.e., $\theta = (k,\lambda)$. Alg.~\ref{alg: quantal level-k dp} shows the procedure to compute the policy of the human induced by his/her latent states. The robot needs to condition the reasoning of human behaviors on the inferred latent states during planning. In addition, the robot's policy and value function as a ql-$k$ agent are also obtained from Alg.~\ref{alg: quantal level-k dp} and are exploited to solve the POMDP in Sec.~\ref{sec: planning algorithm}.
\textbf{Summary.} We model the human's latent states as his/her intelligence level and rationality coefficient, i.e., $\theta = (k,\lambda)$ and use Alg.~\ref{alg: quantal level-k dp} to compute the policies/value functions of the human and the robot as ql-$k$ agents, which are exploited to solve the POMDP in Sec.~\ref{sec: decision-making algorithm}.

\vspace{-0.2cm}
\section{Anytime Active game-theoretic planning}\label{sec: decision-making algorithm}

In this section, we embed the human ql-$k$ model in the POMDP through belief dynamics following the procedure in \cite{Sisi2019,chen2020trust} and present our anytime game-theoretic planner.

\vspace{-0.1cm}
\subsection{Embed the Human Behavioral Model in Robot Planning}\label{sec: embed human model}

\noindent
\textbf{Observation function.} We define an observation made by the robot as $o := \tilde{s}$, i.e., the robot can measure the joint physical state $\tilde{s}$. Then the observation function is defined as: $\mathcal{Z}(o',s) = \mathbb{I}(o' = \tilde{s})$, where $\tilde{s}$ is the joint physical state in $s$, and $\mathbb{I}(\cdot)$ is an indicator function, taking $1$ if the event $(\cdot)$ is true and taking $0$ otherwise.

\noindent
\textbf{Prior belief state.} We define the probability of arriving to state $s'=(\tilde{s}',\theta')\in\mathcal{S}$ from state $s=(\tilde{s},\theta)\in\mathcal{S}$ after executing $a \in\mathcal{A}_{\mathcal{R}}$ as:

\vspace{-.5cm}
\footnotesize
\begin{align}\label{equ: belief transition}
& \mathcal{T}(s,a,s') := \mathbb{P}(s_{t+1} = s' | s_{t} = s, a^{\mathcal{R}}_t = a)\\
%& = \sum_{a' \in \mathcal{A}_{\mathcal{H}}}\mathbb{P}(s_{t+1} = s' | s_{t} = s, a^{\mathcal{R}}_t = a, a^{\mathcal{H}}_t = a')\mathbb{P}(a^{\mathcal{H}}_t = a' | s_{t} = s)\nonumber\\
& =\hspace{-0.2cm} \sum_{a' \in \mathcal{A}_{\mathcal{H}}} \hspace{-0.2cm}\mathbb{I}\big(\tilde{s}' = f(\tilde{s},a,a')\big)\mathbb{P}(\theta_{t+1}=\theta'|\theta_{t}=\theta,\tilde{s}_t=\tilde{s},\bar{\sigma})\pi^{\mathcal{H},k,\lambda}(\tilde{s},a'),\nonumber
\end{align}
\normalsize
\vspace{-0.3cm}

\noindent
where $f$ is the dynamics function described in Sec.~\ref{sec: problem formulation}, $\mathbb{P}(\theta_{t+1}|\theta_{t},\tilde{s}_t,\bar{\sigma})$ represents an explicit probabilistic model that governs the dynamics of the latent states ($\bar{\sigma}$ denotes the model parameters), and $\pi^{\mathcal{H},k,\lambda}$ denotes the human's ql-$k$ policy with rationality coefficient $\lambda$ (recall $\theta=(k,\lambda)$). Then, we can define a prior belief state prediction function that predicts the future belief state without accounting for the possible observations: $\tilde{b}_{t+1} = \tilde{\rho}(b_t, a)$, where each element in $\tilde{b}_{t+1}$ is computed following $ \tilde{b}_{t+1}(s') = \sum_{s\in\mathcal{S}}\mathcal{T}(s,a,s') b_t(s)$. Then, the robot's reward function in belief space can be defined as: $ r_{\mathcal{R}}'(b_t,a_t)= \sum_{\tilde{s}'}r_{\mathcal{R}}(\tilde{s}')\mathbb{P}(\tilde{s}_{t+1}=\tilde{s}'|\tilde{\rho}(b_t,a_t))$.

%\vspace{+0.1cm}
\noindent
\textbf{Prior probability of future observations.} With the observation function and the prior belief prediction function defined, given an initial belief state $b$, we can predict the probabilities of the robot's future observations after executing an action $a\in\mathcal{A}_{\mathcal{R}}$ following:

\vspace{-0.5cm}
\footnotesize
\begin{align}\label{equ: observation prediction}
\mathcal{O}(o,b,a):=&\mathbb{P}(o_{t+1}=o | b_{t} = b, a^{\mathcal{R}}_t = a) = \hspace{-0.2cm}\sum_{s'\in\mathcal{S}} \mathcal{Z}(o, s')\tilde{b}_{t+1}(s').
\end{align}
\normalsize
\vspace{-0.4cm}

\noindent
\textbf{Posterior belief update.} After executing an action and receiving an observation, the robot can update its posterior belief state through the belief dynamics equation $b_{t+1} = \rho(b_t, a^{\mathcal{R}}_t, o_{t+1})$. More specifically, each element in $b_{t+1}$ can be computed using the Bayesian inference equation \cite{sarkka2013bayesian}:

\vspace{-0.5cm}
\small
\begin{align}
    b_{t+1}(s') \propto {\mathcal{Z}(o_{t+1}, s') \tilde{b}_{t+1}(s')}, \quad \forall s' \in \mathcal{S}.
    \label{equ: bayesian}
\end{align}
\normalsize
\vspace{-0.5cm}

\noindent
\textbf{Summary}. The human's behavioral model is embedded into the belief transition \eqref{equ: belief transition}. With \eqref{equ: bayesian}, the robot infers the human's latent states, and in turn uses the inference result to predict belief evolution via the belief prediction function $\tilde{\rho}$.

\vspace{-0.1cm}
\subsection{Robot Planning Algorithm}
\label{sec: planning algorithm}
\vspace{-0.1cm}
\noindent
\textbf{Closed-loop policy.}
In stochastic environments, the robot's actions can help ``actively learn" the latent states for the benefits of the future. \cite{bar1974dual} shows that only the closed-loop policy has the capability of \textit{active learning} as it optimally balances between uncertainty reduction and maximizing utility. Solving \eqref{equ: closed loop policy} via stochastic dynamic programming to obtain a closed-loop policy is computationally intractable \cite{bellman1966dynamic}. To achieve real-time computation, we exploit open-loop strategies, but compensate for the loss of active learning.

%s\vspace{+0.1 cm}
\noindent
\textbf{Open-loop feedback strategy.} %We write the belief space planning problem as a stochastic model predictive control problem.
In contrast to solving for the optimal policy in \eqref{equ: closed loop policy}, we let the robot solve for an optimal action sequence at each $t$:

\vspace{-0.5cm}
\small
\begin{align}
    \mathbf{a}^{\mathcal{R}}_t & = \argmax_{\mathbf{a}} \mathbb{E}_{\mathcal{Z}} \left[ \hat{V}(b_{t+T}) + \sum_{\tau=0}^{T-1} r'_{\mathcal{R}}(b_{t+\tau},a_{t+\tau})\right],
   \label{equ: open-loop optimization}
\end{align}
\normalsize
\vspace{-0.4cm}

\noindent
where $T$ is the planning horizon, $\mathbf{a} = \{a_{t+0},\dots,a_{t+T-1}\}$ is a planned action sequence, and $\hat{V}(b_{t+T})$ denotes the \textit{terminal value} of the predicted belief state $b_{t+T}$. The robot plans in a feed-back manner by applying the first action in $\mathbf{a}_t^{\mathcal{R}}$ and re-planing at the next time step. As opposed to the closed-loop policy, \eqref{equ: open-loop optimization} fixes the action plan ahead and omits the benefits that can be propagated back from future observations. Consequently, \eqref{equ: open-loop optimization} only \textit{passively} learns the latent states and yields conservative actions. Hence, explicit methods can be used to actively learn the latent states \cite{WITTENMARK199567}.

\noindent
\textbf{Active learning of the latent states.} We exploit the Shannon entropy \cite{shannon2001mathematical} to measure the estimation uncertainty of a belief state, and augment the robot's reward function with the expected information gain:

\vspace{-0.47cm}
\small
\begin{align}
\mathcal{I}(b,a) = H(b) - \sum_{o}\mathcal{O}(o,b,a)H\big(\rho(b,a,o)\big),
\end{align}
\normalsize
\vspace{-0.40cm}

\noindent
where \small $H(b) = -\sum_{s\in\mathcal{S}}b(s)\log{b(s)}$\normalsize, and $\mathcal{O}$ is the observation prediction function defined in \eqref{equ: observation prediction}. In general, the higher the $\mathcal{I}(b,a)$ is, the more expected information the robot obtains about the human's latent states if the robot executes $a$ in $b$. With the augmented reward function $\tilde{r}_{\mathcal{R}}(b,a)$, \eqref{equ: open-loop optimization} becomes:

\small
\vspace{-0.5cm}
\begin{subequations}\label{equ: open-loop optimization with safe probing}
\begin{align}
    &\mathbf{a}^{\mathcal{R}}_t  = \argmax_{\mathbf{a}} \mathbb{E}_{\mathcal{Z}} \left[ \hat{V}(b_{t+T})+ \sum_{\tau=0}^{T-1} \tilde{r}_{\mathcal{R}}(b_{t+\tau},a_{t+\tau})\right], \label{eq: action}
\end{align}
\vspace{-0.55cm}
\begin{align}
    \hspace{-2.9cm}\tilde{r}_{\mathcal{R}}(b,a)  =  r'_{\mathcal{R}}(b,a) + \eta\mathcal{I}(b,a),\label{eq: exploration reward}
\end{align}
\vspace{-0.85cm}
\begin{align}
    \hspace{-0.5cm}\text{s.t.} \quad  \mathbb{P}(\tilde{s}_{t+\tau} \in \mathbb{O}_{\text{safe}} | \mathbf{a}, b_t) \geq 1 - \Delta_{\tau}, \sum_{\tau = 1}^{T} \Delta_{\tau} \leq  \Delta,\label{equ: constraints}
\end{align}
\end{subequations}
\normalsize
\vspace{-0.45cm}

\noindent
where  $\eta \propto H(b)$ is an adaptive term that enables the robot to learn the latent states as needed. The constraint \eqref{equ: constraints} requires that the probability that the predicted state $\tilde{s}_{t+\tau}$ is in the safe set $\mathbb{O}_{\text{safe}}$ is larger than $1-\Delta_{\tau}$ for all steps over the planning horizon, with $\Delta_{\tau}$ being a design parameter. The overall safety is bounded by $\Delta$ via risk allocation \cite{blackmore2009convex, ono2012joint} (evaluation of risk is shown in Alg.~\ref{alg: tree search}). \cite{Sisi2019} exploited continuous relaxation techniques to solve \eqref{equ: open-loop optimization} with a time-joint chance constraint. However, with the information reward \eqref{eq: exploration reward}, the approach in \cite{Sisi2019} could become intractable.

\vspace{-0.2cm}
\begin{algorithm}
    \footnotesize
    \caption{\small Open-Loop Monte-Carlo Belief Tree Search}
    \label{alg: tree search}
    
    \textbf{Input}: POMDP model, planning horizon $T$, discount factor $\gamma$, exploration parameter $e$, and initial belief state $b_0$;
    
    \While {goal not reached}{
    %Initialize $\mathcal{V}$,  and $N$;
     Initialize root node $v_{\text{root}} = (\emptyset, 0, 0)$;
     
    \While{not TimeOut()}{
        Simulate($v_{\text{root}}$, $b_t$, $0$);
        $N(v_{\text{root}}.\mathbf{a}) += 1 $; 
    }
    
    $a^{\mathcal{R}}_t = \argmax_{a \in \mathcal{A}_{\mathcal{R}}}V(a)$;
    
    Execute $a^{\mathcal{R}}_t$ and collect the new observation $o_{t+1}$;
    
    Update belief state: $b_{t} \leftarrow \rho(b_t,a_t,o_{t+1})$;
    }
    \texttt{\\}
    % Simulate Function
    \textbf{Function} Simulate($v$, $b$, $\tau$)
    
    % reach the end, retern the terminal cost
    \If{$\tau = T-1$}{return $\sum_{k\in\mathcal{K}}\Big( \mathbb{P}(k^{\mathcal{H}}=k|b)\big(\sum_{\tilde{s} \in \tilde{\mathcal{S}}} \mathbb{P}( \tilde{s}|b)V^{*,\mathcal{R},k+1}(\tilde{s})\big)\Big)$}
    \eIf{ IsLeafNode($v$)}{
        \For{all a $\in$ $\mathcal{A}_{\mathcal{R}}$}{
           % StageRisk = ComputeRisk$(\tilde{\rho}(\tau,a))$.
            
            \If{ComputeRisk$(\tilde{\rho}(b,a))$ $<\Delta_{\tau}$}
            % Add the next action in the tree
            {
             
            Append $v_{succ} = \big<v.\mathbf{a}.add(a),0,0\big>$ in the tree;
            }

        }
        
        return \textit{Rollout}($b$, $\tau$);
    }{
    % Tree policy
        $v^*_{\text{succ}} = \argmax_{v_{\text{succ}}} V(v_{\text{succ}}.\mathbf{a}) + e\sqrt{\frac{\log(N(v.\mathbf{a}))}{N(v_{\text{succ}}.\mathbf{a})}}$;
        
        $a = v^*_{\text{succ}}.\mathbf{a}.last()$; Sample observation $o \sim \mathcal{O}(o,b,a)$;
        
        $ \hat{V} = \tilde{r}_{\mathcal{R}}(b,a) + \gamma$Simulate($v^*_{\text{succ}}$, $\rho(b,a,o)$, $\tau+1$);

        %$N(v.\mathbf{a}) \leftarrow N(v.\mathbf{a}) + 1 $;
        $N(v^*_{\text{succ}}.\mathbf{a}) += 1 $;
        $V(v^*_{\text{succ}}.\mathbf{a}) += \frac{ \hat{V} - V(v^*_{\text{succ}}.\mathbf{a})}{N(v^*_{\text{succ}}.\mathbf{a})} $; Return $\hat{V}$;
    }

    \textbf{end Function}

    \texttt{\\}
    % Time-Joint chance constraint evaluation
    \textbf{Function} \textbf{ComputeRisk}(b)
    
     return $\sum_{\theta\in \Theta}\text{proj}_{\theta}(b)^{\intercal} [\mathbb{I}(\tilde{s}^1 \in \mathbb{O}^c_{\text{safe}}), \dots, \mathbb{I}(\tilde{s}^{n_{\tilde{s}}} \in \mathbb{O}^c_{\text{safe}})]^{\intercal}$;
    
    \textbf{end Function}
    
    %\texttt{\\}
    % Compute terminal value
    
    % \textbf{Function}\textbf{ ComputeTerminalValue ($b$)}
    
    %  return $\sum_{k\in\mathcal{K}} \mathbb{P}(k^{\mathcal{H}}=k|b)\sum_{\tilde{s} \in \tilde{\mathcal{S}}} \mathbb{P}(s = \tilde{s}|b)V^{*,k+1,1}(\tilde{s})$
    
    % \textbf{end Function}
    
     \texttt{\\}
    % Rollout
    
    \textbf{Function}\textbf{ Rollout ($b$, $\tau$)}
    
     \eIf{$\tau = T - 1$}{
       return $\sum_{k\in\mathcal{K}}\Big( \mathbb{P}(k^{\mathcal{H}}=k|b)\big(\sum_{\tilde{s} \in \tilde{\mathcal{S}}} \mathbb{P}( \tilde{s}|b)V^{*,\mathcal{R},k+1}(\tilde{s})\big)\Big)$
     }
     {
        $a\sim\pi_{rollout}(b)$;$o\sim\mathcal{O}(o,b,a)$;
    
        return $\tilde{r}_{\mathcal{R}}(b,a) + \gamma\cdot${Rollout}($\rho(b,a,o)$, $\tau+1$);
     }
    
    \textbf{end Function}
    
    \normalsize
\end{algorithm}
\vspace{-0.3cm}

%We note that with the chance-constraints imposed, the approach that exploits the gradient based optimization method (L-BFGS) in \cite{Sadighinformation} can no longer be applied.

\noindent
\textbf{Open-loop chance-constrained Monte-Carlo belief tree search.} POMCP \cite{silver2010monte} combines  Monte-Carlo simulation and  game-tree search. Building upon POMCP, we propose an algorithm (Alg. \ref{alg: tree search}) to solve \eqref{equ: open-loop optimization with safe probing} in an anytime manner. Our algorithm differs from POMCP in the following ways: 1) a search node, $v = \big< \mathbf{a}, V(\mathbf{a}), N(\mathbf{a}) \big>$, stores an action sequence $\mathbf{a}$ and its associated statistics: $V(\mathbf{a})$ is the mean return of all simulations that execute $\mathbf{a}$, and $N(\mathbf{a})$ counts the number of times that $\mathbf{a}$ has been visited; 2) leaf node expansions must enforce the safety chance constraint (line 14); 3) terminal values are estimated using the pre-computed ql-$k$ values (line 11); 4) the active information gathering on latent states is explicitly realized via the augmented reward function $\tilde{r}_{\mathcal{R}}$.

\noindent
\textbf{Benefits of \cref{alg: tree search}.} A search node only stores an action sequence since the open-loop optimization \eqref{equ: open-loop optimization with safe probing} searches for an action sequence. In contrast to POMCP, in which a node stores a history of actions and observations, the search space in Alg.~\ref{alg: tree search} is significantly reduced. Hence Alg.~\ref{alg: tree search} can run in real-time under computational constraints. Quantal level-$k$ reasoning (Sec.~\ref{sec: internal state modeling}) is exploited to estimate the terminal value when the maximum planning depth is reached (line 11). Specifically, the terminal belief state $b_{t+T}$ is used to determine the human's intelligence level distribution $\mathbb{P}(k^{\mathcal{H}}|b_{t+T})$, then, we assume the robot behaves as a ql-$(k^{\mathcal{H}}+1)$ agent (recall that a ql-$(k+1)$ agent quantally best responds to a ql-$k$ agent) with rationality coefficient $\lambda = 1$ and estimates the terminal value using the pre-computed robot's ql-$(k^{\mathcal{H}}+1)$ value weighted by $\mathbb{P}(k^{\mathcal{H}}|b_{t+T})$. By actively learning the belief state, the planner can quickly reduce the estimation error on the terminal values and maximize its performance.

\noindent
\textbf{Infeasibility handling mechanism.} When the root node has no safe successors, we relax the chance constraint and find the action that minimizes the degree of constraint violation.%: $a^*_{\mathcal{R}} = \argmin_{a} ComputeRisk(\tilde{\rho}(b_t,a))$. 

\section{Experiments}

\subsection{Implementation Details}\label{sec: implementation details}

\begin{figure}[h]
\begin{center}
\begin{picture}(300.0, 40)
\put(8, 5){\epsfig{file=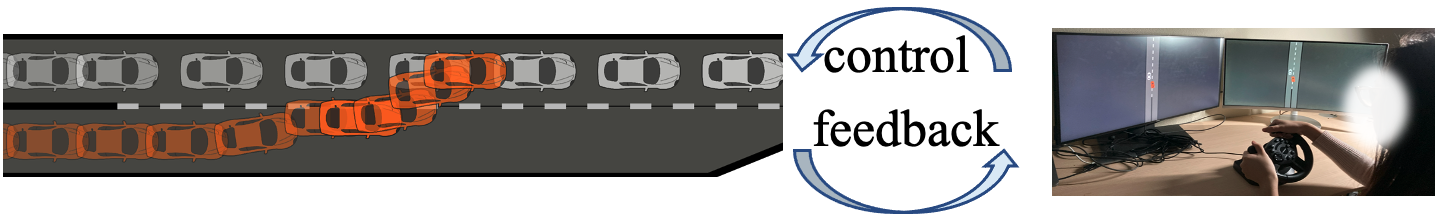,width = 0.9 \linewidth,angle=0, trim=0cm 0.0cm 0.0cm 0.0cm,clip}}
\small
\normalsize
\end{picture}
\end{center}
\vspace{-0.8cm}
\caption{\hspace{-0.3cm} An autonomous car interacts with a human-driven car controlled by a user through a racing wheel and a pedal in a forced merging scenario.}
\label{fig: intro_fig}
\vspace{-0.2cm}
\end{figure}

\noindent
\textbf{Test domain.} We use \textit{autonomous driving} as the test domain. In particular, we consider a forced merging scenario \cite{isele2019interactive}, where an autonomous car must merge to an adjacent (upper) lane that is occupied by a human-driven car (\cref{fig: intro_fig}).

\noindent
\textbf{Reward functions.} The reward function of a car is a linear combination of features \small$\phi: \tilde{S}\rightarrow \mathbb{R}^{n_f}$,\normalsize  satisfying $r_{(\cdot)}(\tilde{s}) = \omega^\intercal \cdot\phi(\tilde{s})$. The features encode safety, comfort, progress, etc. \cite{tian2018adaptive}. The weights $\omega \in \mathbb{R}^{n_f}$ can be recovered via an inverse learning algorithm \cite{ziebart2008maximum, sun2019interpretable, tian2020bounded}. Note that when the autonomous car runs Alg. \ref{alg: tree search}, the safety feature in its reward function is deactivated since the safety is handled by \eqref{equ: constraints}.

\noindent
\textbf{Level-0 policy.} Alg.~\ref{alg: quantal level-k dp} requires a ql-$0$ policy to initiate the iterative reasoning process. Similar to \cite{LiUnsignalized, tian2018adaptive}, we let a ql-$0$ agent be a non-strategic agent who treats others as static obstacles when making decisions.% Note that this ql-$0$ policy is chosen specifically for the autonomous driving setting, and it can differ according to the application.

\noindent
\textbf{Latent state dynamics.} Recall \eqref{equ: belief transition} that an explicit probabilistic model is required to govern the dynamics of the latent states. In this work, we consider single-shot games, i.e., the human's latent states are assumed to be constant during interaction. Hence, the latent states' transition model is reduced to $\mathbb{P}(\theta_{t+1}|\theta_{t},\tilde{s}_t,\bar{\sigma}) {=} \mathbb{I}(\theta_{t{+}1}{=}\theta_t)$. In general repeated games, the transition model can be represented as a Markov chain and its parameters $\bar{\sigma}$ can be embedded in the POMDP and learned simultaneously as in \cite{tianbeating}.

\noindent
\textbf{High-level robot planning.} The dynamics of the human-robot team are represented as

\small
\vspace{-0.5cm}
\begin{align}
    \hspace{-0.3cm}\left[ \begin{matrix}\dot{x}_{\mathcal{R}} & \dot{y}_{\mathcal{R}} & \dot{x}_{\mathcal{H}} & \dot{v}_{\mathcal{R}} & \dot{v}_{\mathcal{H}}\end{matrix}\right] = \left[\begin{matrix}v_{\mathcal{R}} & w_{\mathcal{R}} & v_{\mathcal{H}} & a_{\mathcal{R}} & a_{\mathcal{H}}\end{matrix} \right],\label{equ: unicycle model}
\end{align}
\vspace{-0.5cm}
\normalsize

\noindent where $x$ ($y$) is the longitudinal (lateral) position, $v$ ($w$) is the longitudinal (lateral) speed, and $a$ is the acceleration. The sampling period is $\Delta t = 0.5[s]$. We use a state grid of the size $40\times6\times40\times6\times6$ to represent the discrete states of the human-robot system.
The safety set $\mathbb{O}_{\text{safe}}$ includes states in which the boundaries of the two agents do not overlap. We use $\Delta {=} 0.05$ and $\Delta_{\tau} {=} \frac{1}{160}$ as the chance constraint thresholds in \eqref{equ: constraints}. We let the highest intelligence level of the human be $k_{\text{max}}{=}2$ based on experimental results in \cite{costa2006cognition,costa2009comparing}. The rationality coefficients take value from the set $\Lambda{=}\{0.5, 0.8, 1.0\}$. The planning horizon in \eqref{equ: open-loop optimization with safe probing} is $T {=} 8$.

\noindent
\textbf{Hierarchical planning and control.} 
Behavioral planning and control of the autonomous car are hierarchically connected. The planning layer (Alg.~\ref{alg: tree search}) uses a low-fidelity model \eqref{equ: unicycle model} to generate behavioral commands, and runs at $8Hz$ on a laptop with 2.8 GHz CPU. In the low-level control layer (running at $8Hz$), the vehicle dynamics are represented by a high-fidelity bicycle model \cite{kong2015kinematic}, and we use a model predictive controller \cite{borrelli2017predictive, cvxgen} to generate continuous controls that drive the system to the desired states generated by Alg.~\ref{alg: tree search}. In Alg.~\ref{alg: tree search}, when the actual system state deviates from the state grid, the nearest neighboring grid will be exploited.

\noindent
\textbf{Baselines.} We consider two baseline planners: 1) our planner without the feature of active information gathering, i.e., the autonomous car passively infers the human's latent states (BLP-1); 2) the strategic game-theoretic planner in \cite{fisac2019hierarchical} that treats the human-driven car as a \textit{follower} who accommodates the actions from the autonomous car (BLP-2). Both BLP-2 and our planner use a closed-loop feedback structure when building human behavioral models, but our planner reasons about the heterogeneity in the human's cognitive limitations and irrationality through active inference rather than treating the human as a follower.

\vspace{-0.1cm}
\subsection{The Human Behavioral Model}
\label{sec: quantal level-k illuetration}

\noindent
\textbf{Human intelligence level interpretation.} The ql-$k$ model is exploited to reason about human behaviors under bounded intelligence. Recall that the level-$0$ agent represents a non-strategic agent who treats others as static obstacles. Thus, a ql-$1$ agent can be interpreted as a cautious agent since it believes that its opponent is an aggressive non-strategic agent. On the contrary, a ql-$2$ agent behaves aggressively since it believes its opponent is a cautious ql-$1$ agent.

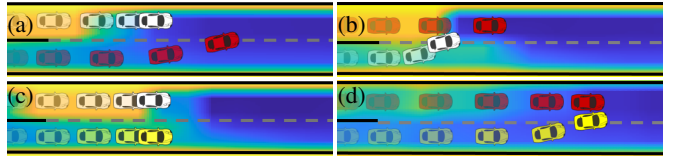
\begin{figure}[ht]
\begin{center}
\begin{picture}(200.0, 50)
\put(-25, 30){\epsfig{file=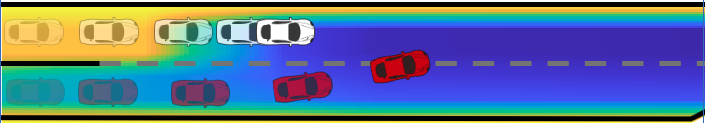,width = 0.5 \linewidth, height = 1.0cm,angle=0, trim=1cm 0.0cm 4cm 0.0cm,clip}}
\put(100, 30){\epsfig{file=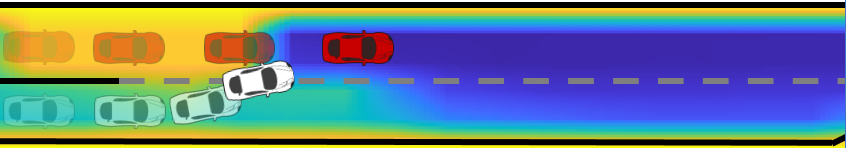,width = 0.5 \linewidth, height = 1.0cm, angle=0, trim=1cm 0.0cm 4cm 0.0cm,clip}}
\put(-25, 0){\epsfig{file=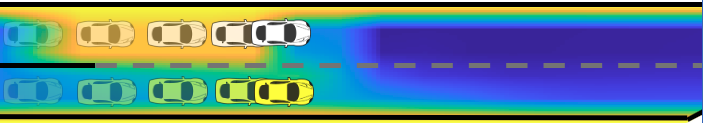,width = 0.5 \linewidth, height = 1.0cm,angle=0, trim=1cm 0.0cm 4cm 0.0cm,clip}}
\put(100, 0){\epsfig{file=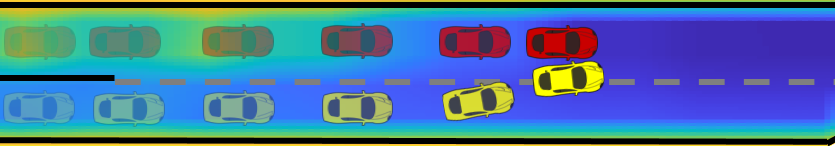,width = 0.5 \linewidth, height = 1.0cm, angle=0, trim=1cm 0.0cm 4cm 0.0cm,clip}}
\small
\put(-25,20){(c)}
\put(-25,47){(a)}
\put(100,20){(d)}
\put(100,47){(b)}
\normalsize
\end{picture}
\end{center}
\vspace{-0.5cm}
\caption{\textbf{Interactions between ql-$k$ agents}. (a-b): ql-1 (white) v.s. ql-2 (red); (c): ql-1 v.s. ql-1; (d): ql-2 v.s. ql-2. All agent have $\lambda =1$, the same initial longitudinal position, and the same initial speed $12 [m/s]$.}
\label{fig: ql-k illustration - optimal}
\end{figure}

\cref{fig: ql-k illustration - optimal} shows the interactions between two cars modeled as ql-$k$ agents in the forced merging task. The heat-map displays the ql-$k$ value function ($V^{*,i,k}$ described in Sec.~\ref{sec: human qlk reasoning}) of the lower-lane car, indicating the preferred states; colder color means higher value. It can be observed in (a-b) that the interactions are seamless between a ql-$1$ agent and a ql-$2$ agent, which is expected since the ql-$2$ agent's belief in the model of the ql-$1$ agent matches the ground truth (note that the high-value region in the upper lane encourages the red car to merge ahead of the white car (a); the low-value region in front of the white car guides it to yield (b)). However, when an agent's true model deviates from the other agent's belief, conflicts may occur. For instance, (c) shows the interaction between two ql-$1$ agents with a dead-lock because both agents prefer to yield. The low-value region in front of the yellow car discourages it to merge although it is safe. Similarly, when two ql-$2$ agents interact (d), collisions may occur as both agents think their opponents will be likely to yield. Note that the high value region in the upper lane encourages the yellow car to merge even if it is not safe.

\vspace{-0.3cm}

% \di{I'm not sure what you mean here by projected.}
% \tr{About the value map projection, note that the value function is a function of $\tilde{s}$, the physical state of the human-robot team, thus for a pair of ($\tilde{s}_1$, $\tilde{s}_2$), we can have one value point. The projection of the value on one upper car's state means: if we fix the upper lane car's state (e.g., $\tilde{s}_1 = (0,0,0)$), then we can have a value map, representing the value of the lower car in any state on the grid, and showing which regions are preferred if the upper lane car's state is $\tilde{s}_1$. The projection of the value on one upper car's trajectory is just a average of the projection of the values on each state in the trajectory. I agree it's a little confusing, so I deleted the  projection part just saying its the ql-k value, do you think that's ok?}

\vspace{-0.3cm}
\subsection{Case Studies and Quantitative Results}
\label{sec: case studies}

We compare our planner against the baselines via simulations. The human driven-car is modeled as a ql-$k$ agent, and is also controlled by the hierarchical planning and control scheme described in Sec.~\ref{sec: implementation details}, with behavioral commands obtained directly from the corresponding ql-$k$ policies.

\noindent\textbf{Hypotheses.} We state the following two hypotheses: 1). active exploration improves efficiency of the robot's planning; 2). our planner is robust to the human's heterogeneous intelligence levels and irrational behaviors.

%%%%%%%%%%%% Case study I %%%%%%%%%%%%
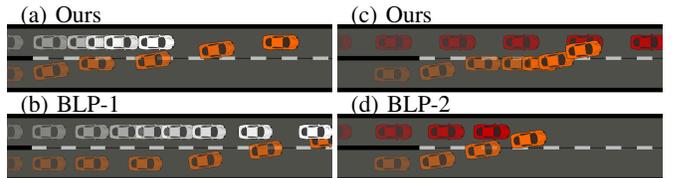
\begin{figure}[ht]
\begin{center}
\begin{picture}(200.0, 65)
\put(-25, 34){\epsfig{file=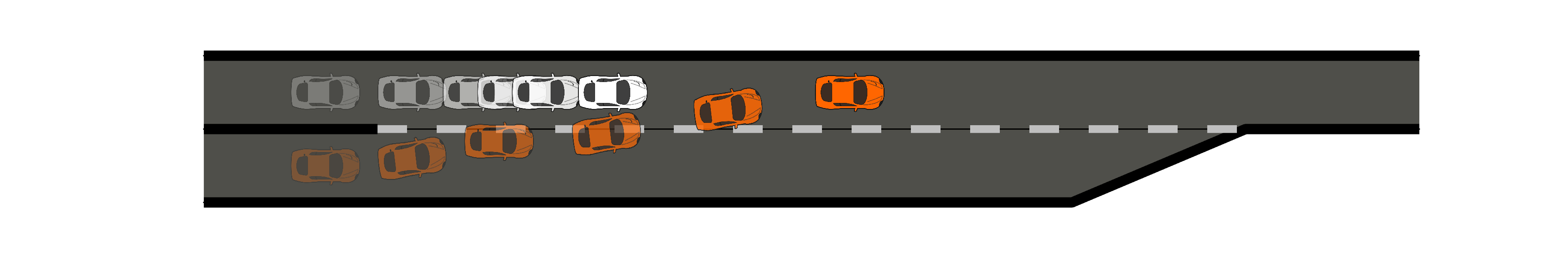,width = 0.5 \linewidth, height = 1.25cm,angle=0, trim=4.5cm 0.0cm 8.5cm 0.5cm,clip}}
\put(-25, 0){\epsfig{file=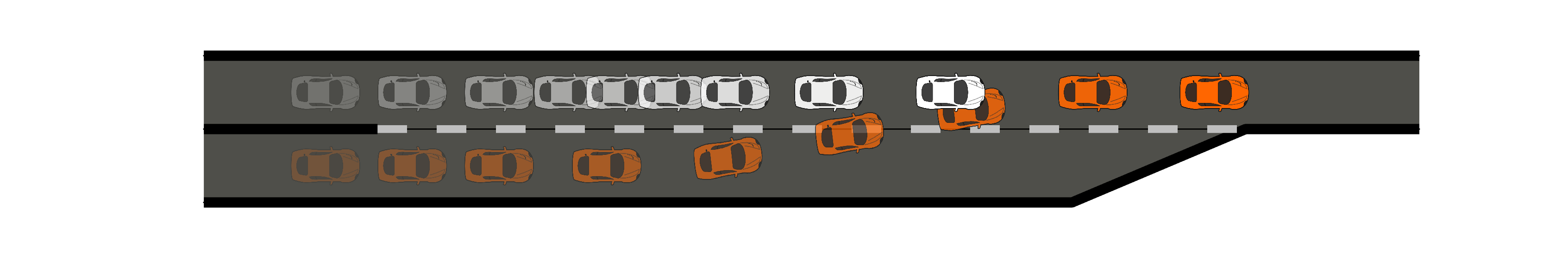,width = 0.5 \linewidth, height = 1.25cm,angle=0, trim=4.5cm 0.0cm 8cm 0.5cm,clip}}
%\put(-25, 0){\epsfig{file=media/level_2_car_with_cautious_human.png,width = 0.5 \linewidth, height = 1.25cm,angle=0, trim=5cm 0.0cm 8cm 0.5cm,clip}}

\put(100, 34){\epsfig{file=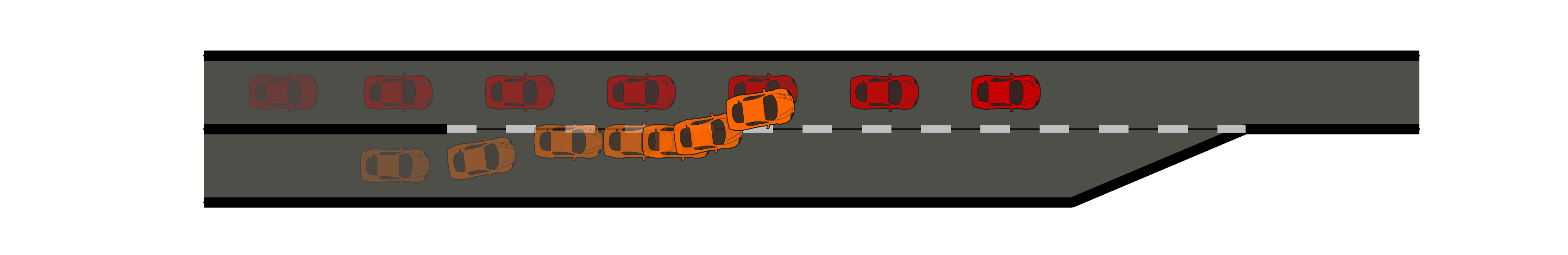,width = 0.5 \linewidth, height = 1.25cm,angle=0, trim=4cm 0.0cm 9cm 0.5cm,clip}}
%\put(100, 32){\epsfig{file=media/normal_car_with_aggressive_human_ahead.png,width = 0.5 \linewidth, height = 1.25cm,angle=0, trim=4cm 0.0cm 9cm 0.5cm,clip}}
\put(100, 0){\epsfig{file=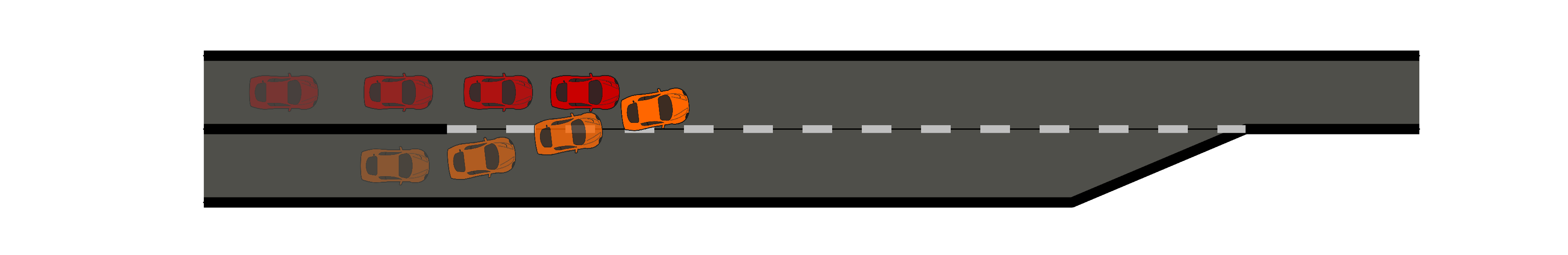,width = 0.5 \linewidth, height = 1.25cm,angle=0, trim=4cm 0.0cm 9cm 0.5cm,clip}}
\small
\put(-20,68){(a) Ours}
\put(-20,34){(b) BLP-1}
%\put(-20,34){(c) BLP-2}
\put(105,68){(c) Ours}
%\put(105,66){(e) BLP-1}
\put(105,34){(d) BLP-2}
%\put(25,43){\footnotesize \textit{\orange{nudging in to explore}}}
%\put(150,43){\footnotesize \textit{\orange{nudging in to explore}}}
\normalsize
\end{picture}
\end{center}
\vspace{-0.8cm}
\caption{\textbf{Planner comparison.} (a-b) show the interactions between a simulated \textit{cautious} human-driven car (white) and the autonomous car (orange); (c-d) show those between a simulated \textit{aggressive} human-driven car (red) and the autonomous car (orange). Initial speeds are $12 ~[m/s]$.} 
\label{fig: scenario-1}
\vspace{-0.4cm}
\end{figure}

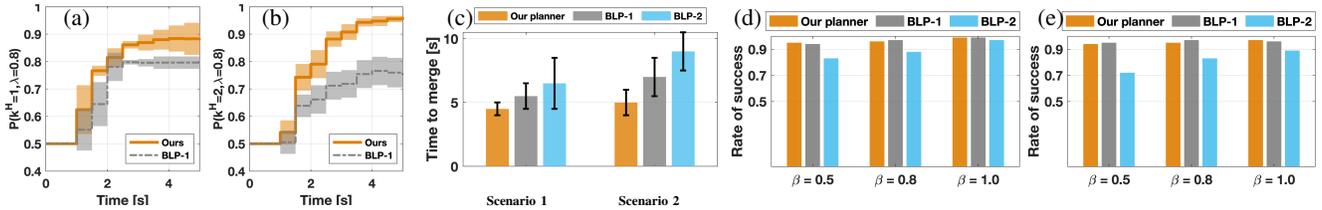
\begin{figure*}[thp!]
\centering
\begin{picture}(200.0, 75.0)
\put(-52,  0){\epsfig{file=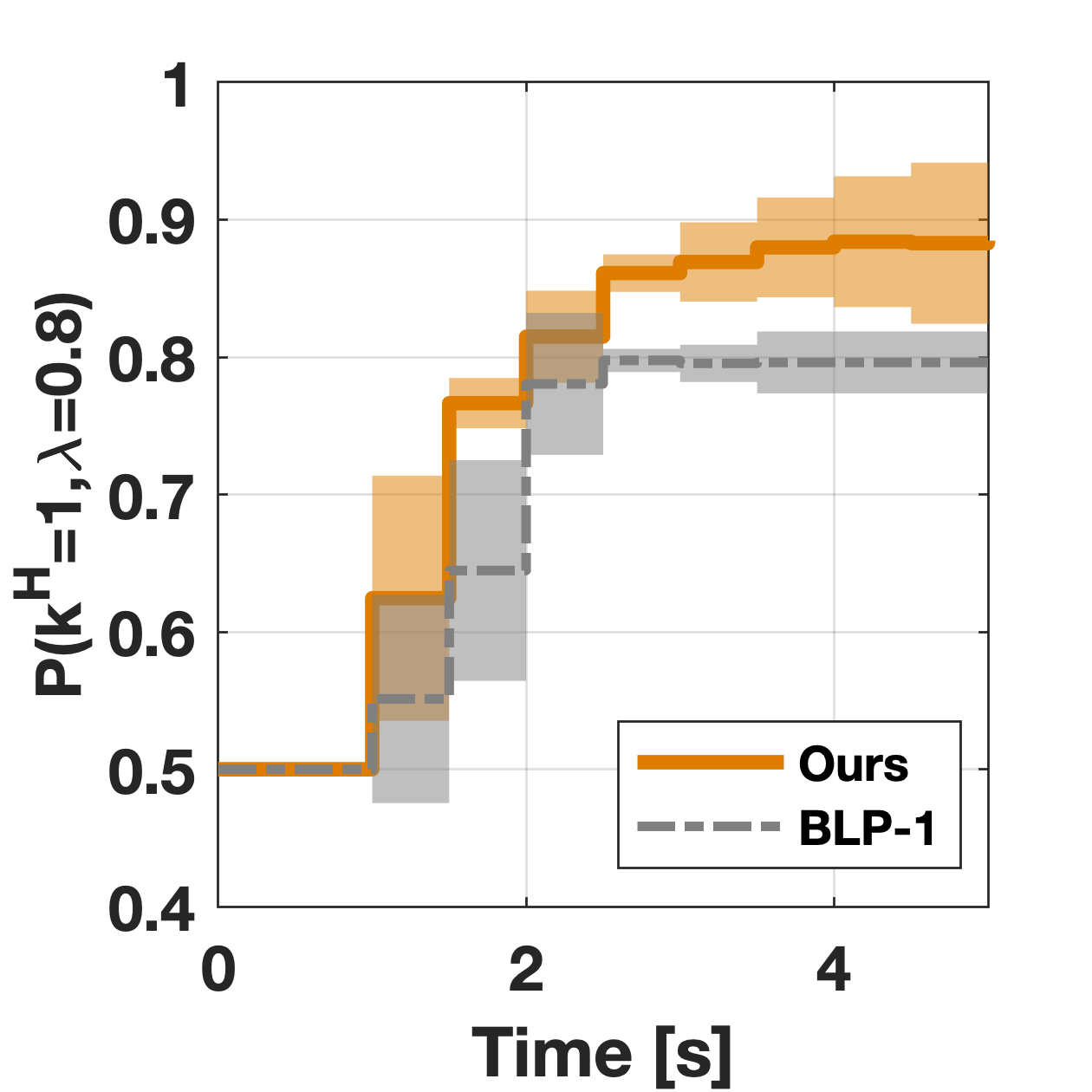,width = 0.16 \linewidth, trim=0.1cm 0.1cm 0cm 0.5cm,clip}}
\put(25,  0){\epsfig{file=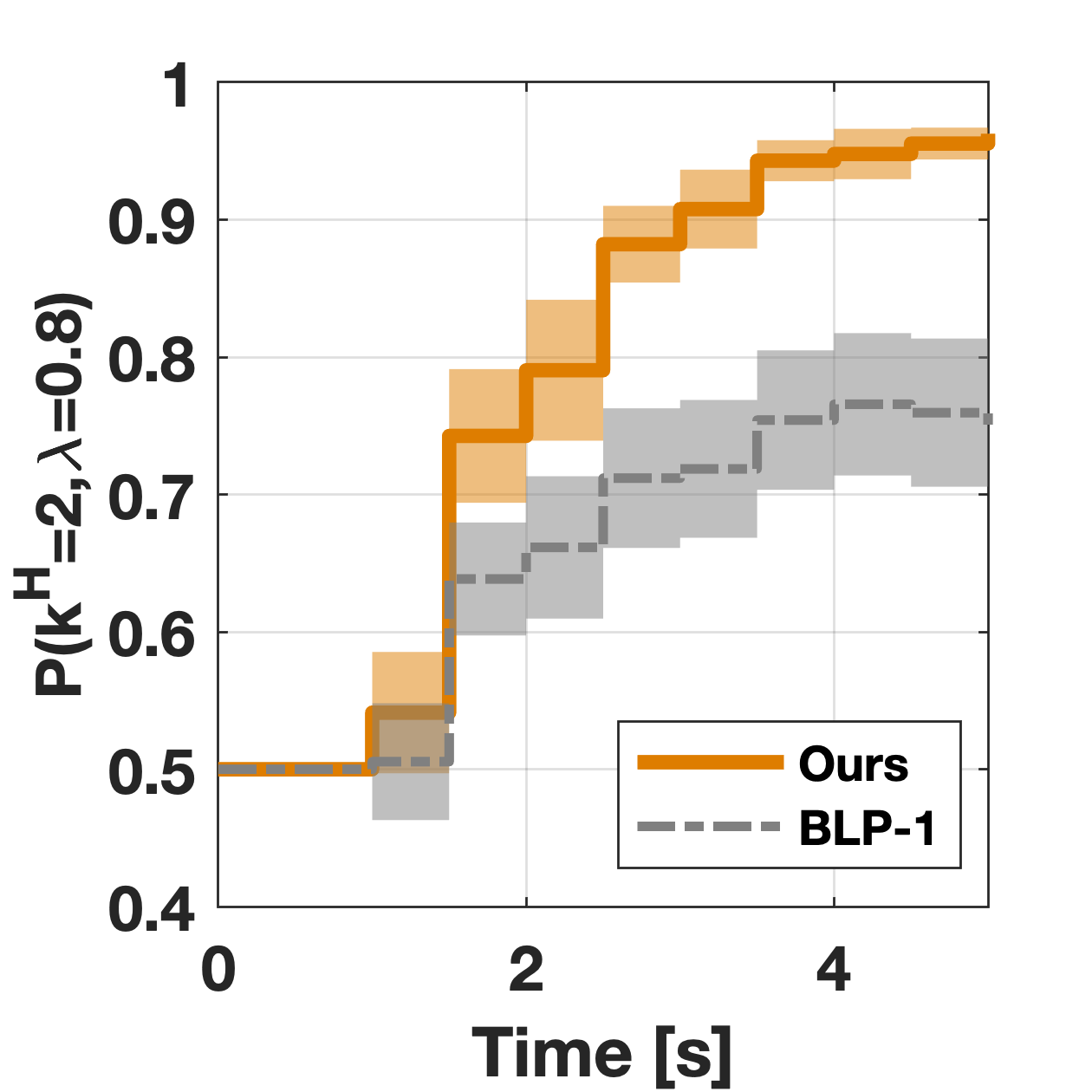,width = 0.16 \linewidth, trim=0.1cm 0.1cm 0cm 0.5cm,clip}}
\put(105, 12){\epsfig{file=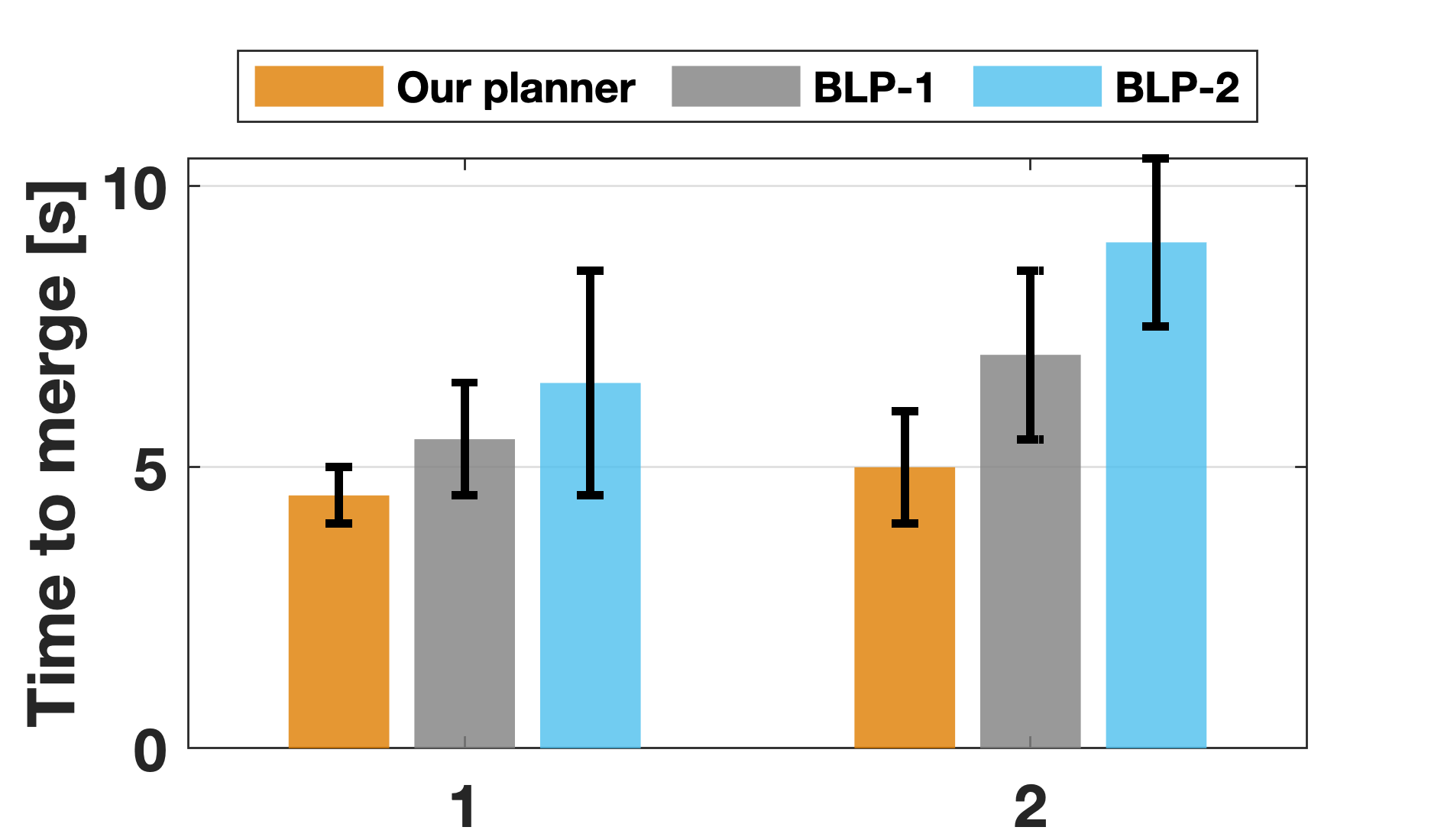,width = 0.245 \linewidth,angle=0, trim=0cm 0.5cm 0cm 0.3cm,clip}}
\small
\put(-30,68){(a)}
\put(45,68){(b)}
\put(115,68){(c)}
\put(130,0){\tiny\textbf{Scenario 1}}
\put(180,0){\tiny\textbf{Scenario 2}}
\normalsize
\end{picture}
\begin{picture}(200.0, 75.0)
\put(  18,  7){\epsfig{file=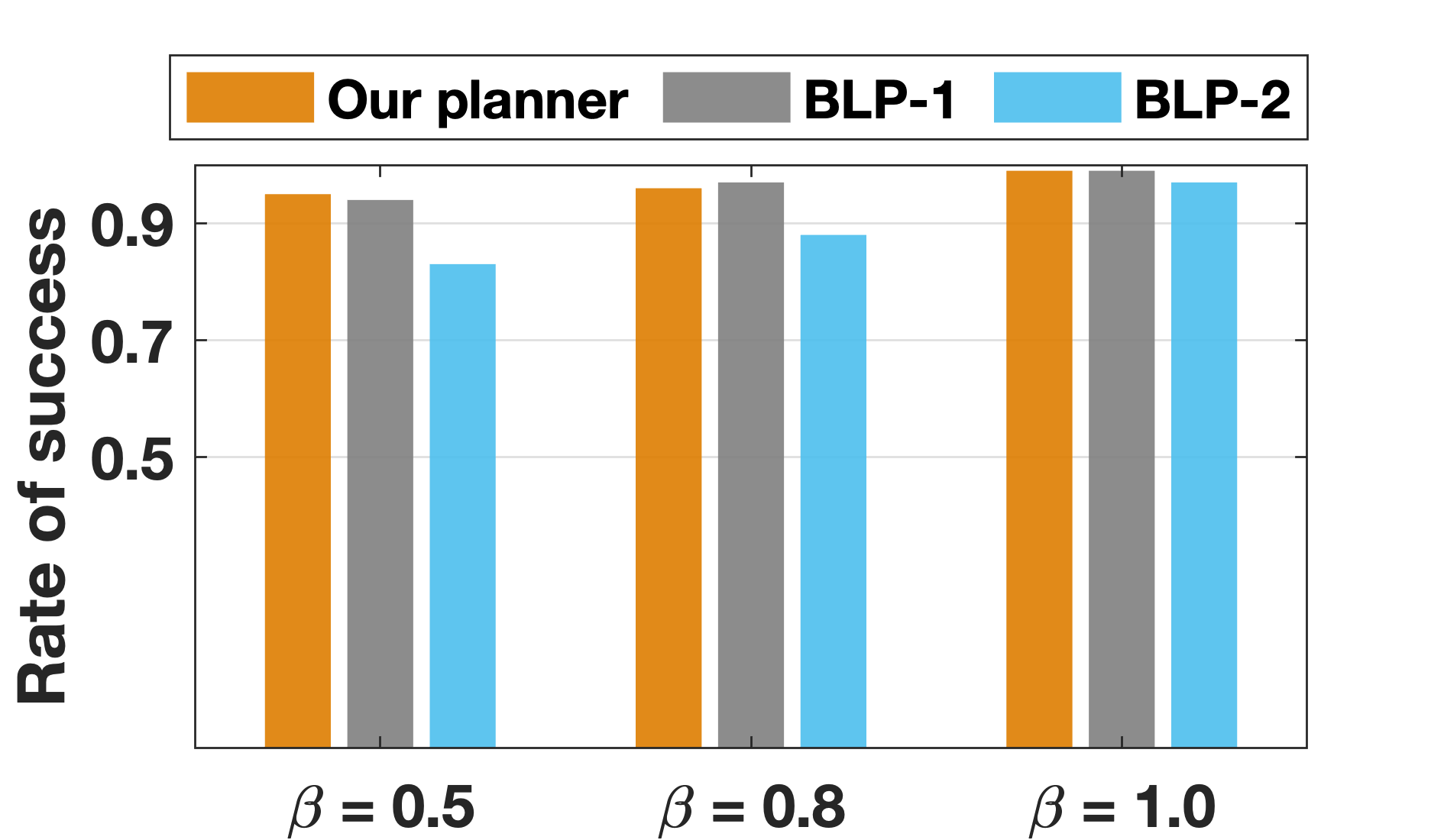,width = 0.24 \linewidth, trim=0.0cm 0.0cm 0cm 0.3cm,clip}}  %%%
%%%%%%%%%%%%
\put(  130,  7){\epsfig{file=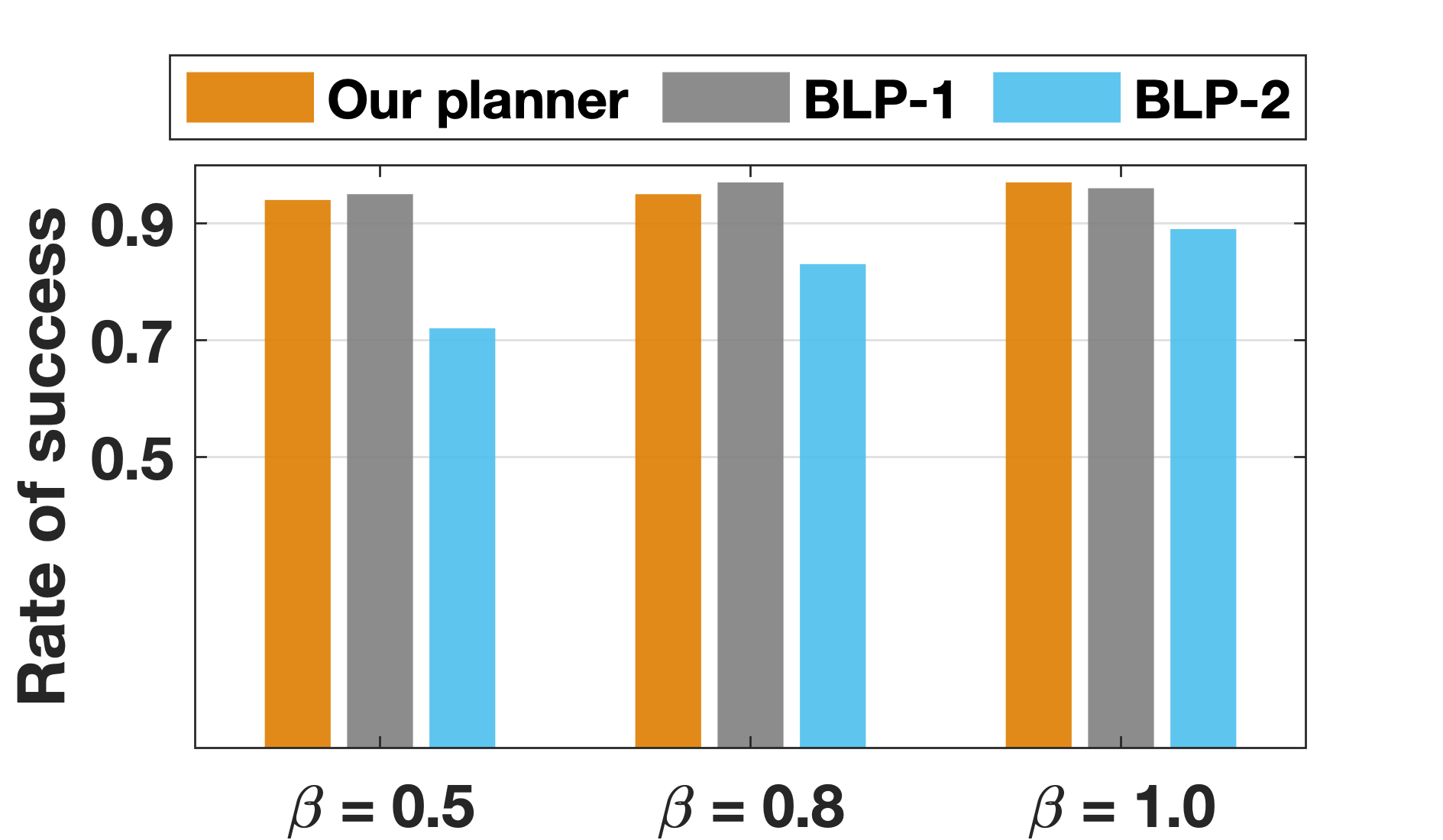,width = 0.24 \linewidth, trim=0.0cm 0.0cm 0cm 0.3cm,clip}}  %%%
%%%%%%%%%%%%%%%%%%%%%%
\small
\put(20,68){(d)}
\put(135,68){(e)}
\normalsize
\end{picture}
\vspace{-0.3cm}
  \caption{\textbf{Evaluation result.} (a-b): Human latent states inference results for (a) cautious and (b) aggressive humans. (c): Average TM of the autonomous car; error bars show 95\% confidence integral. (d-e): The rates of success under various human rationality coefficients (higher $\lambda$ yields more rational human behaviors). The human-driven car is a simulated cautious driver with intelligence level-$1$ in (d), and an aggressive driver with intelligence level-$2$ in (e).}
  \label{fig: sim-success}
\vspace{-0.5cm}
\end{figure*}

We first show two case studies, highlighting the benefits that naturally emerged from our planner.

\noindent \textbf{Scenario 1.} Fig.~\ref{fig: scenario-1}(a-b) show a scenario where the autonomous car and a simulated cautious human-driven car (ql-$1$ agent) with rationality coefficient $\lambda=0.8$ start at the same initial speed and longitudinal position. It can be observed that with our planner, the autonomous car actively indicates an intention to merge by nudging into the target lane, as it predicts that, the human's \textit{reactions} triggered by the ``probing" action can help disambiguate the human's latent states. With the baseline planner (BLP-1), the autonomous car takes a longer time to merge, since passive inference requires additional observations to infer the latent states.\enspace

\noindent \textbf{Scenario 2.} \cref{fig: scenario-1}(c-d) show a case that is similar to scenario 1 but the human-driven car is simulated by an aggressive ql-$2$ agent who starts behind the autonomous car. With our planner, the autonomous car nudges in to explore the human's latent states, then quickly decides to yield after observing the humans' aggressive reactions. With BLP-2, the autonomous car initiates a dangerous merge as BLP-2 incorrectly assumes the human-driven car will likely yield. Note that when humans behave under heterogeneous cognitive states, it is critical for a robot to reason about such a heterogeneity to better predict human behaviors.

\noindent
\textbf{Metrics for quantitative results.} In quantitative studies, we use two metrics for validating the hypotheses : 1) the rate of success (RS), which measures the percentage of simulations in which no collision or dead-lock occurs; 2) the time used by the autonomous car to complete the merge (TM).

\noindent
\textbf{Results.} We run $50$ simulations for each of the scenarios in the case studies using our planner and two baseline planners. We first compare our planner against BLP-1 in terms of inference performance (BLP-2 has no inference capability). In \cref{fig: sim-success} (a-b), we show the time history of the autonomous car's belief in the ground truth human-driven car's latent states. It can be observed that our planner can more effectively identify the human's latent states by exploiting the mutual influence to make its human partner reveal his/her hidden states. In \cref{fig: sim-success} (c), we show the average TM for the two scenarios. Note that our planner achieves the lowest TM, due to the active inference. {Furthermore, our planner achieves the highest confidence on TM since our planner strategically generates actions that aim for triggering the most informative reactions from the human, this regulates human behaviors in a game-theoretic sense (such a phenomena is also observed in user studies \cref{fig: human-experiment}(b)).}

% FIGURE 4
% \begin{figure}[ht]
% \begin{center}
% \begin{picture}(200.0, 60.0)
% \put(-26,  0){\epsfig{file=media/cautious_human_inference_result.png,width = 0.3 \linewidth, trim=0.1cm 0.1cm 0cm 0.5cm,clip}}
% \put(45,  0){\epsfig{file=media/aggressive_human_inference_result.png,width = 0.3 \linewidth, trim=0.1cm 0.1cm 0cm 0.5cm,clip}}
% \put(120, 10){\epsfig{file=media/TTM_compare.png,width = 0.4 \linewidth,angle=0, trim=0cm 0.0cm 1cm 0.0cm,clip}}
% \small
% \put(-10,62){(a)}
% \put(61,62){(b)}
% \put(121,62){(c)}
% \put(150,5){\tiny\textbf{Scenario}}
% \normalsize
% \end{picture}
% \end{center}
% \vspace{-0.5cm}
%       \caption{(a-b): Human latent state inference results. (c): Average TM of the autonomous car; error bar means 95\% confidence integral.}
%       \label{fig: sim-eval}
% \end{figure}
% \vspace{-0.3cm}

Next, we evaluate the robustness of our planner under various human intelligence levels and degrees of irrationality. We let the simulated ql-$k$ human take his/her rationality coefficient from $\Lambda$ and run $100$ simulations for each ($\lambda, k$) combination. The human starts at a random position within 10 $m$ around the autonomous car. In \cref{fig: sim-success}(d-e), we show the RS of each planner for each scenario. Note that in all cases, our planner and BLP-1 achieve more than 95\% RS. This is attributed to reasoning about the human's bounded intelligence and irrationality, and enforcing the safety chance constraint. BLP-2 shows satisfactory RS when the simulated human is a ql-$1$ agent, but the RS decreases as the human becomes more irrational. When the human is a ql-$2$ agent, BLP-2 performs noticeably worse. The results indicate that our planner enables the robot to learn the human's latent states, plan more effectively, and be robust to the human's various intelligence levels and irrational behaviors.

\subsection{User Study}
\label{sec: user study}

\noindent
\textbf{Objective.} We conduct user studies in which we let the autonomous car interact with a real human driver in a simulator (\cref{fig: intro_fig}), showcasing the effectiveness of our planner.

\noindent
\textbf{Experiment setup.} We recruited 10 human participants. For each participant, we ran Scenario 1 for $3$ times using each of the planners. Here, the accelerations of the upper lane car are provided by human participants directly through a pedal. 

\begin{figure}[ht]
\begin{center}
\begin{picture}(200.0, 56)
%%%%%%%%%%%%%%%%%%%%%%%%%%%%%%
\put(-23, 0){\epsfig{file=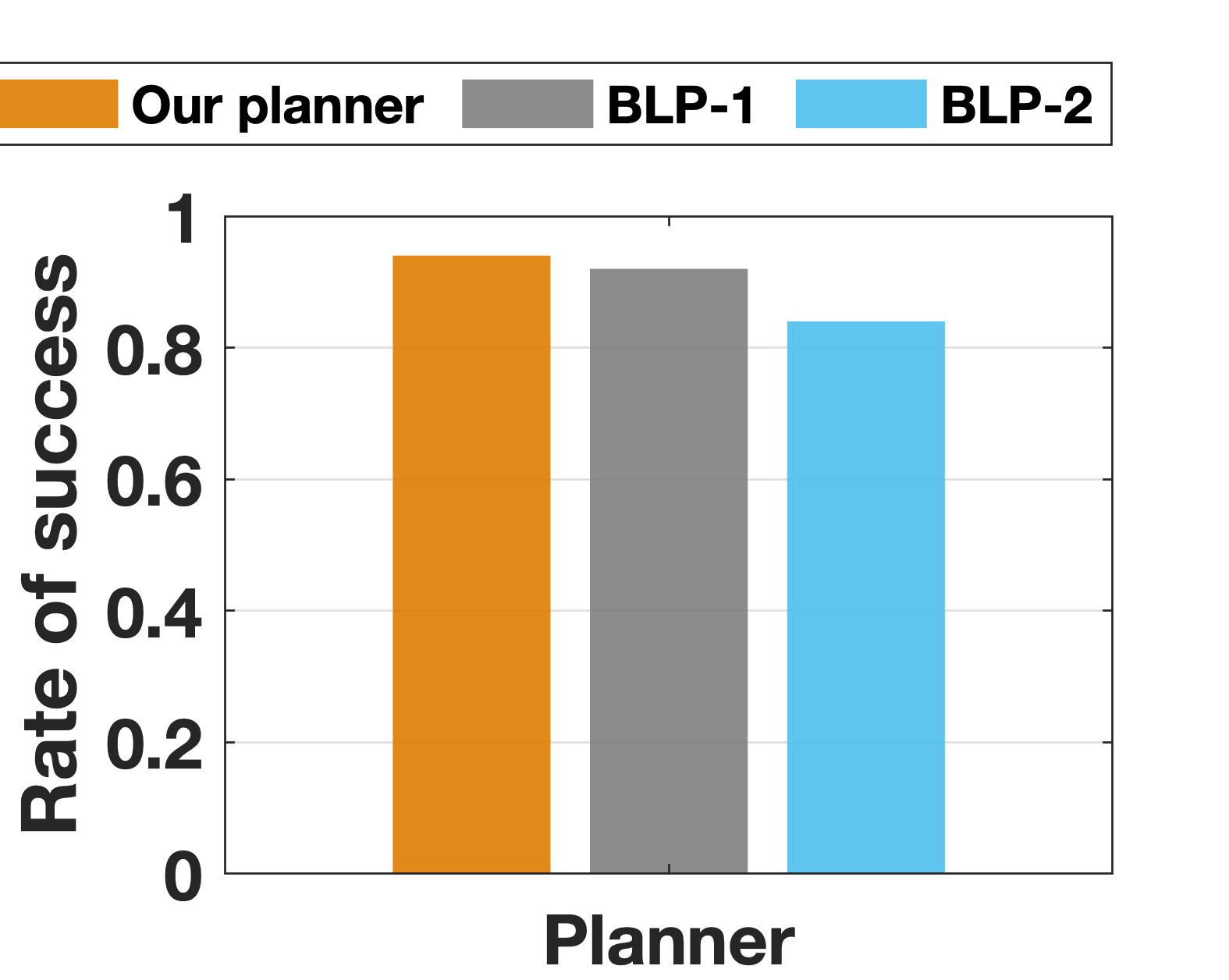,width = 0.33 \linewidth,angle=0, trim=0cm 0cm 0.0cm 0.0cm,clip}}
\put(53, -2){\epsfig{file=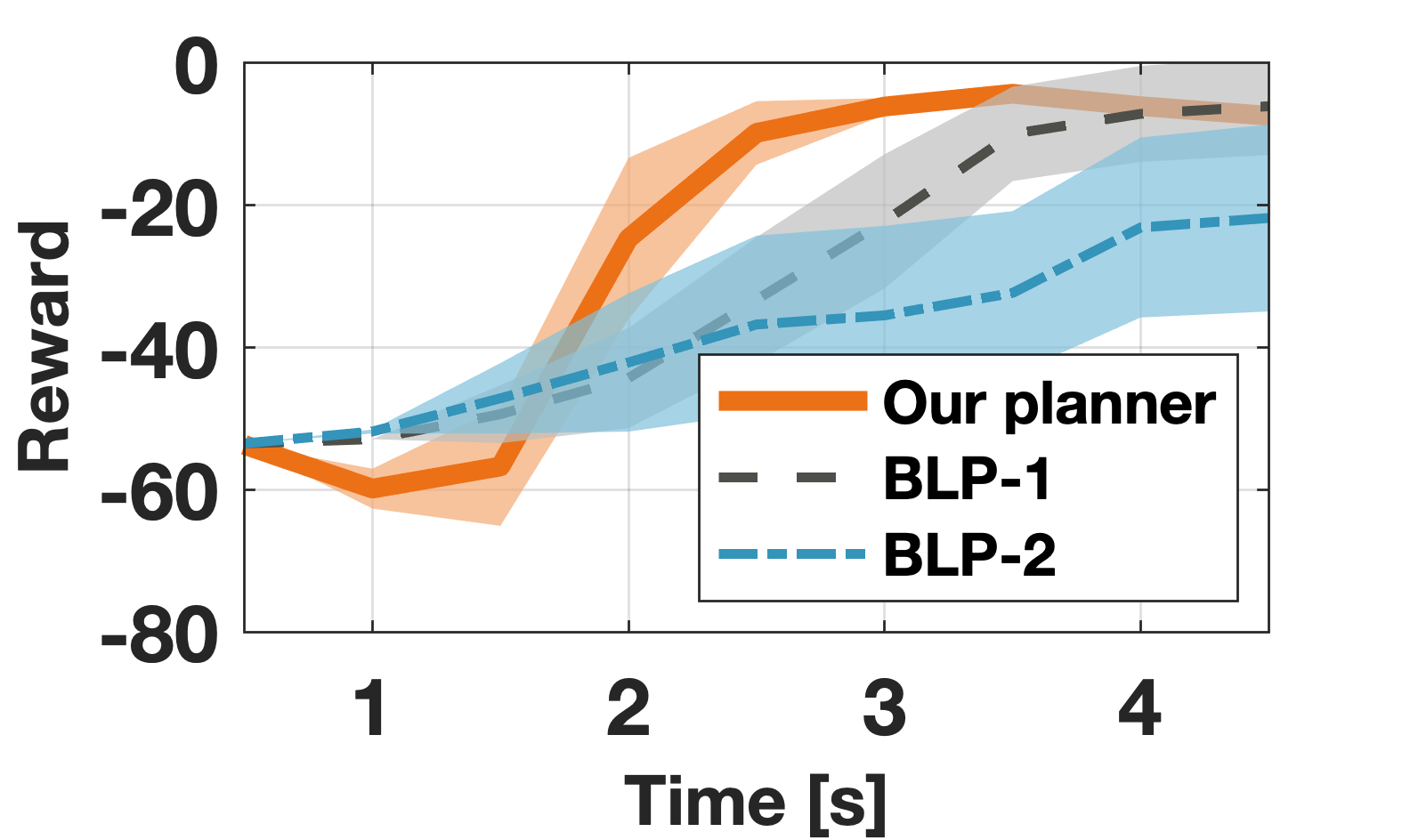,width = 0.4 \linewidth,angle=0, trim=0cm 0.0cm 0.5cm 0.0cm,clip}}
\put(150, -5){\epsfig{file=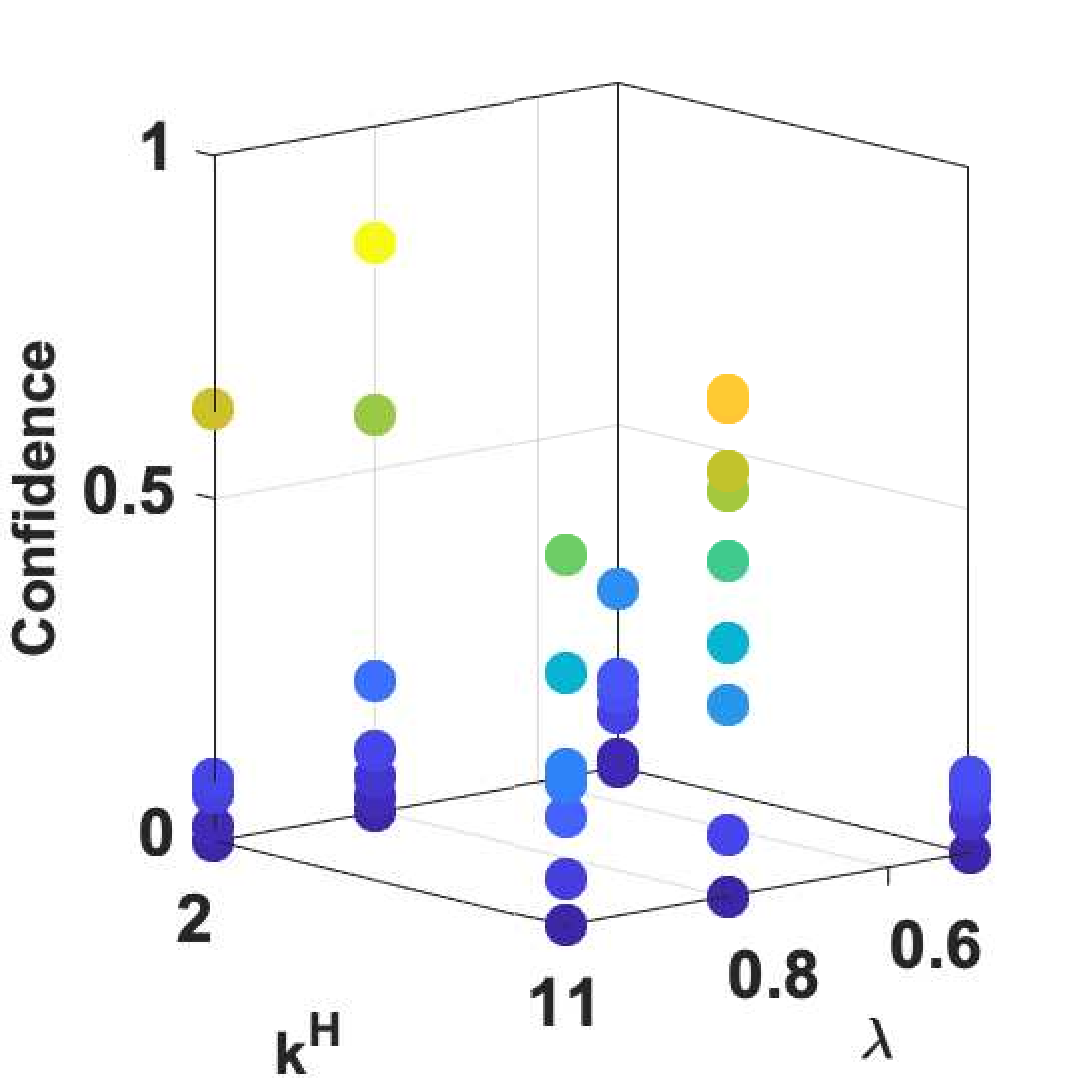,width = 0.29 \linewidth,angle=0, trim=0.cm 0.0cm 0.0cm 0.5cm,clip}}
\small
\normalsize
\end{picture}
\end{center}
\vspace{-0.5cm}
\caption{\textbf{User study results}: (a) the rates of success of the planners; (b) the time histories of the step reward of the autonomous car; the thick line represents the mean and the shaded area represents the 95\% confidence tube of the data; (c) the distribution of inferred intelligence levels and rationality coefficients of the human participants.}
\label{fig: human-experiment}
\vspace{-0.3cm}
\end{figure}

\noindent
\textbf{Results.} \cref{fig: human-experiment}(a) shows the RS of each planner. It can be observed that both our planner and BLP-1 outperform the BLP-2 in terms of safety. \cref{fig: human-experiment}(b) shows the time histories of the step reward collected by the autonomous car. Note that although the autonomous car pays more costs in the first few steps using our planner (due to the probing actions) it is able to complete the task much more effectively and safely compared with the baselines. In \cref{fig: human-experiment}(c), we show the distributions of inferred intelligence levels and rationality coefficients. It can be observed that roughly 60\% of the participants are identified as ql-$1$ agents, which is aligned with the experiment study on other forms of games  \cite{costa2009comparing}. In addition, \cref{fig: human-experiment}(c) suggests that, under the reward function assumed by the autonomous car, most of the participants demonstrated behaviors that align with the behaviors produced by the rationality coefficient $\lambda=0.8$. % (i.e., less rational behaviors).
\vspace{-0.1cm}
\section{Conclusion}

We proposed an anytime game-theoretic planning framework that integrates iterative reasoning models, POMDP, and chance-constrained Monte-Carlo belief tree search for robot behavioral planning. Our planner enables a robot to safely and actively reason about its human partner’s latent cognitive states in real-time to maximize its utility more effectively. We applied the proposed approach to an autonomous driving domain where our behavioral planner and a low-level motion controller hierarchically control an autonomous car to negotiate traffic merges. Both simulation and user study results demonstrated the effectiveness of our planner compared with baseline planners. %Our future work focuses on applying the proposed framework in repeated games that have more complex latent states dynamics. 

%\floatbarrier
\newpage

% \noindent\textbf{Conclusion}

%\begin{IEEEkeywords}

%\end{IEEEkeywords}

% \fi
% creates the second title. It wipassivekyll be ignored for other modes.
\IEEEpeerreviewmaketitle

\vspace{-0.3cm}
%\begin{small}
\setstretch{0.98}
\bibliographystyle{IEEEtran}

%%\bibliography{ref} C:\Users\hp\Documents
\bibliography{ICRA_2020_Ref}
%\end{small}

%\section{Appendix}

%\input{treesearch_algo}

\end{document}